\documentclass[sigconf,screen]{acmart}
\usepackage{latexsym}
\usepackage{url}
\usepackage{xcolor}
\definecolor{newcolor}{rgb}{.8,.349,.1}
\usepackage{times}
\usepackage{epsfig}
\usepackage{graphicx}
\usepackage{amsmath}
\usepackage{graphicx}
\usepackage{booktabs}
\usepackage{enumitem}
\usepackage{colortbl}
\usepackage{float}
\usepackage{algpseudocode}
\usepackage[ruled]{algorithm2e}
\usepackage{caption}
\usepackage{subcaption}
\usepackage{multirow}
\usepackage{enumitem}
\usepackage{etoolbox}
\usepackage[normalem]{ulem}
\usepackage{csquotes}
\usepackage{dblfloatfix}
\definecolor{mygray}{gray}{.9}
\definecolor{ForestGreen}{RGB}{34,139,34}
\newcommand\ie{i.e.}

\usepackage{tcolorbox}
\newtcolorbox[list inside=prompt,auto counter,number within=section]{prompt}[1][]{
    colbacktitle=black!60,
    coltitle=white,
    fontupper=\footnotesize,
    boxsep=5pt,
    left=0pt,
    right=0pt,
    top=0pt,
    bottom=0pt,
    boxrule=1pt,
    title={#1},
    #1, 
}

\newcommand\dataset{AV-Deepfake1M}
\newcommand\datasetabbr{AV-Deepfake1M}

\AtBeginDocument{%
  \providecommand\BibTeX{{%
    \normalfont B\kern-0.5em{\scshape i\kern-0.25em b}\kern-0.8em\TeX}}}

\setcopyright{acmlicensed}
\copyrightyear{2024}
\acmYear{2024}
\setcopyright{rightsretained}
\acmConference[MM '24]{Proceedings of the 32nd ACM International Conference on Multimedia}{October 28-November 1, 2024}{Melbourne, VIC, Australia}
\acmBooktitle{Proceedings of the 32nd ACM International Conference on Multimedia (MM '24), October 28-November 1, 2024, Melbourne, VIC, Australia}
\acmDOI{10.1145/3664647.3680795}
\acmISBN{979-8-4007-0686-8/24/10}

\acmSubmissionID{0005}

\begin{document}

\title{AV-Deepfake1M: A Large-Scale LLM-Driven Audio-Visual Deepfake Dataset}



\author{Zhixi Cai}
\orcid{0000-0001-7978-0860}
\affiliation{%
  \institution{Monash University}
  \city{Melbourne}
  \country{Australia}}
\email{zhixi.cai@monash.edu}

\author{Shreya Ghosh}
\orcid{0000-0002-2639-8374}
\affiliation{%
  \institution{Curtin University}
  \city{Perth}
  \country{Australia}}
\email{shreya.ghosh@curtin.edu.au}

\author{Aman Pankaj Adatia}
\orcid{0009-0007-9585-404X}
\affiliation{%
  \institution{Indian Institute of Technology Ropar}
  \city{Ropar}
  \country{India}}
\email{2020csb1154@iitrpr.ac.in}

\author{Munawar Hayat}
\orcid{0000-0002-2706-5985}
\affiliation{%
  \institution{Qualcomm}
  \city{San Diego}
  \country{United States}}
\email{hayat@qti.qualcomm.com}

\author{Abhinav Dhall}
\orcid{0000-0002-2230-1440}
\affiliation{%
  \institution{Flinders University}
  \city{Adelaide}
  \country{Australia}}
\email{abhinav.dhall@flinders.edu.au}

\author{Tom Gedeon}
\orcid{0000-0001-8356-4909}
\affiliation{%
  \institution{Curtin University}
  \city{Perth}
  \country{Australia}}
\email{tom.gedeon@curtin.edu.au}

\author{Kalin Stefanov}
\orcid{0000-0002-0861-8660}
\affiliation{%
  \institution{Monash University}
  \city{Melbourne}
  \country{Australia}}
\email{kalin.stefanov@monash.edu}

\renewcommand{\shortauthors}{Cai et al.}

\begin{abstract}
The detection and localization of highly realistic deepfake audio-visual content are challenging even for the most advanced state-of-the-art methods. While most of the research efforts in this domain are focused on detecting high-quality deepfake images and videos, only a few works address the problem of the localization of small segments of audio-visual manipulations embedded in real videos. In this research, we emulate the process of such content generation and propose the AV-Deepfake1M dataset. The dataset contains content-driven (i) video manipulations, (ii) audio manipulations, and (iii) audio-visual manipulations for more than 2K subjects resulting in a total of more than 1M videos. The paper provides a thorough description of the proposed data generation pipeline accompanied by a rigorous analysis of the quality of the generated data. The comprehensive benchmark of the proposed dataset utilizing state-of-the-art deepfake detection and localization methods indicates a significant drop in performance compared to previous datasets. The proposed dataset will play a vital role in building the next-generation deepfake localization methods. The dataset and associated code are available at \href{https://github.com/ControlNet/AV-Deepfake1M}{https://github.com/ControlNet/AV-Deepfake1M}.

\begin{table*}[t]
\centering
\caption{\textbf{Details for publicly available deepfake datasets in a chronologically ascending order.} \textmd{Cla: Binary classification, SL: Spatial localization, TL: Temporal localization, FS: Face swapping, RE: Face reenactment, TTS: Text-to-speech, VC: Voice conversion.}}\vspace{-2mm}
\label{tab:datasets}
\scalebox{0.85}{
\begin{tabular}{l||c|c|c|c|c|c|c|c}
\toprule[0.4mm]
\rowcolor{mygray}\textbf{Dataset} & \textbf{Year} & \textbf{Tasks} & \textbf{Manipulated} & \textbf{Manipulation} & \textbf{\#Subjects} & \textbf{\#Real} & \textbf{\#Fake} & \textbf{\#Total} \\
\rowcolor{mygray}&  &  & \textbf{Modality} & \textbf{Method} &  &  &  & \\
\hline\hline
DF-TIMIT~\cite{korshunovDeepFakes2018} & 2018 & Cla & V & FS & 43 & 320 & 640 & 960 \\
UADFV~\cite{yangExposing2019} & 2019 & Cla & V & FS & 49 & 49 & 49 & 98  \\
FaceForensics++~\cite{rosslerFaceForensics2019} & 2019 & Cla & V & FS/RE & - & 1,000 & 4,000 & 5,000 \\
Google DFD~\cite{nickContributing2019} & 2019 & Cla & V & FS & 5 & 363 & 3,068 & 3,431 \\
DFDC~\cite{dolhanskyDeepFake2020} & 2020 & Cla & AV & FS & 960 & 23,654 & 104,500 & 128,154 \\
DeeperForensics~\cite{jiangDeeperForensics12020} & 2020 & Cla & V & FS & 100 & 50,000 & 10,000 & 60,000 \\
Celeb-DF~\cite{liCelebDF2020} & 2020 & Cla & V & FS & 59 & 590 & 5,639 & 6,229 \\
WildDeepfake~\cite{ziWildDeepfake2020} & 2020 & Cla & - & - & - & 3,805 & 3,509 & 7,314 \\
FFIW$_{10K}$~\cite{zhouFace2021} & 2021 & Cla/SL & V & FS & - & 10,000 & 10,000 & 20,000 \\
KoDF~\cite{kwonKoDF2021} & 2021 & Cla & V & FS/RE & 403 & 62,166 & 175,776 & 237,942 \\
FakeAVCeleb~\cite{khalidFakeAVCeleb2021} & 2021 & Cla & AV & RE & 600$+$ & 570 & 25,000$+$ & 25,500$+$ \\
ForgeryNet~\cite{heForgeryNet2021} & 2021 & SL/TL/Cla & V & Random FS/RE & 5,400$+$ & 99,630 & 121,617 & 221,247 \\
ASVSpoof2021DF~\cite{liuASVspoof2023} & 2021 & Cla & A & TTS/VC & 160 & 20,637 & 572,616 & 593,253 \\
LAV-DF~\cite{caiYou2022} & 2022 & TL/Cla & AV & Content-driven RE/TTS & 153 & 36,431 & 99,873 & 136,304 \\
DF-Platter~\cite{narayanDFPlatter2023} & 2023 & Cla & V & FS & 454 & 133,260 & 132,496 & 265,756 \\
\hline
\dataset{} & 2023 & TL/Cla & AV & Content-driven RE/TTS & 2,068 & 286,721 & 860,039 & 1,146,760 \\
\bottomrule[0.4mm]
\end{tabular}}
\end{table*}
\end{abstract}


\begin{teaserfigure}
  \includegraphics[width=\textwidth]{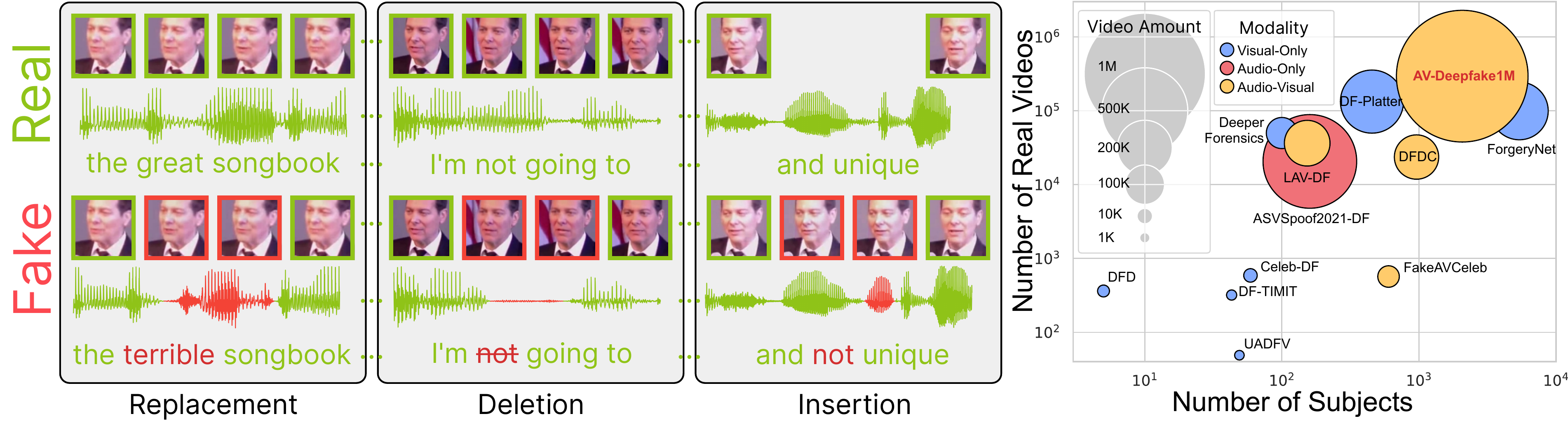}
  \caption{\textbf{\datasetabbr{} is a large-scale content-driven deepfake dataset generated by utilising a large language model.} \textmd{The dataset contains more than 2K subjects and 1M deepfake videos generated by employing different audio-visual content manipulation strategies. The left figure illustrates examples of word-level \emph{replacement}, \emph{deletion}, and \emph{insertion} to manipulate audio-visual content. The right figure provides a comparison between \datasetabbr{} and other publicly available datasets in terms of number of subjects, and amount of real and fake videos.}}
\label{fig:teaser}
\vspace{2mm}
\end{teaserfigure}



\maketitle

\section{Introduction}
\label{sec:intro}

We are witnessing rapid progress in the domain of content generation technology, i.e., models trained on massive amounts of data that can produce highly realistic text~\cite{brownLanguage2020, touvronLLaMA2023}, video~\cite{singerMakeAVideo2022, geLong2022, wuTuneAVideo2023} and audio~\cite{shenNaturalSpeech2023, jiangMegaTTS2023, jiangMegaTTS2023a}.
Consequently, discriminating between real and fake content is becoming increasingly more challenging even for humans~\cite{zhouFace2021, narayanDFPlatter2023}.
This opens the door for misuse of content generation technology for example to spread misinformation and commit fraud, rendering the development of reliable detection methods vital.

The development of such methods is highly dependent on the available deepfake benchmark datasets, which led to the steady increase in the number of publicly available datasets that provide examples of visual-only~\cite{jiangDeeperForensics12020, liCelebDF2020, kwonKoDF2021}, audio-only~\cite{liuASVspoof2023, yiADD2022}, and audio-visual~\cite{khalidFakeAVCeleb2021} content modification strategies (e.g., face-swapping, face-reenactment, etc.).
However, the majority of these datasets and methods assume that the entirety of the content (i.e., audio, visual, audio-visual) is either real or fake.
This leaves the door open for criminals to exploit the embedding of small segments of manipulations in the otherwise real content.
As argued in~\cite{caiYou2022}, this type of targeted manipulation can lead to drastic changes in the underlying meaning as illustrated in Figure~\ref{fig:teaser}.
Given that most deepfake benchmark datasets do not include this type of partial manipulations, detection methods might fail to perform reliably on this new type of deepfake content.

This work addresses this gap by releasing a new large-scale audio-visual dataset called \datasetabbr{} specifically designed for the task of temporal deepfake localization.
To improve the realism and quality of generated content, the proposed data generation pipeline incorporates the ChatGPT\footnote{\url{https://chat.openai.com/}} large language model.
The pipeline further utilizes the latest open-source state-of-the-art methods for high-quality audio~\cite{casanovaYourTTS2022, kimConditional2021} and video~\cite{wangSeeing2023} generation.
The scale and novel modification strategies position the proposed dataset as the most comprehensive audio-visual benchmark as illustrated in Figure~\ref{fig:teaser}, making it an important asset for building the next generation of deepfake localization methods.
The main contributions are,

\begin{itemize}
\item{We propose \datasetabbr{}, a large content-driven audio-visual dataset for the task of temporal deepfake localization.}
\item{We propose an elaborate data generation pipeline employing novel manipulation strategies and incorporating the state-of-the-art in text, video and audio generation.}
\item{We perform comprehensive analysis and benchmark of the proposed dataset utilizing state-of-the-art deepfake detection and localization methods.}
\end{itemize}

\section{Related Work}
\label{sec:related_work}

\begin{figure*}[t]
\centering
\includegraphics[width=\textwidth]{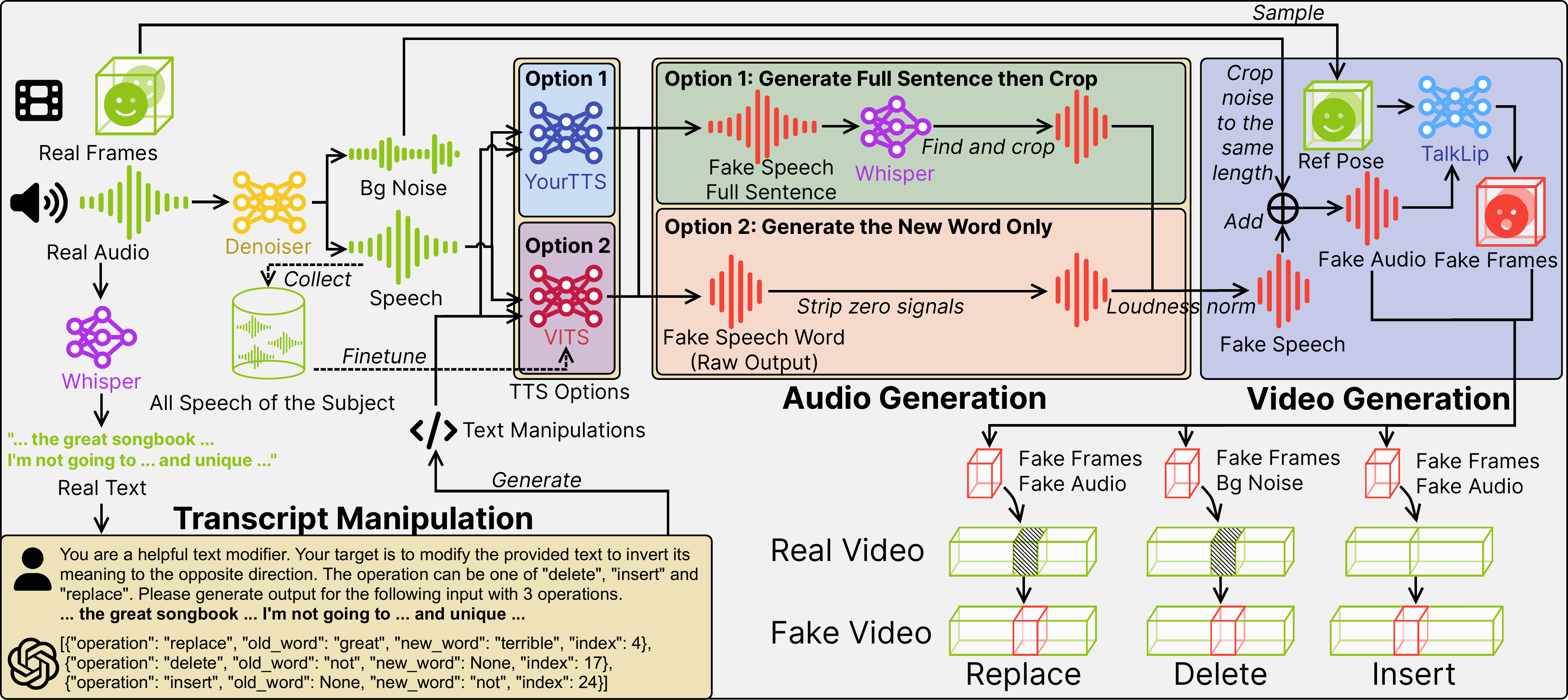}\vspace{-2mm}
\caption{\textbf{Data manipulation and generation pipeline.} \textmd{Overview of the proposed three-stage pipeline. Given a real video, the pre-processing consists of audio extraction via FFmpeg followed by Whisper-based transcript generation. In the first stage, transcript manipulation, the original transcript is modified through word-level insertions, deletions, and replacements. In the second stage, audio generation, based on the relevant transcript manipulation, the audio is generated in both speaker-dependent and independent fashion. In the final stage, video generation, based on the generated audio, the subject-dependant video is generated with smooth transitions in terms of lip-synchronization, pose, and other relevant attributes.}}
\label{fig:pipeline}
\vspace{-3mm}
\end{figure*}

The performance of any deepfake detection method is highly dependent on the quantitative and qualitative aspects of the datasets used for development~\cite{zhangUMMAFormer2023, wangM2TR2022, shaoDetecting2024, wangExploiting2024, oorloffAVFF2024, yangAVoiDDF2023, yuPVASSMDD2023, ilyasAVFakeNet2023, fengSelfSupervised2023, razaMultimodaltrace2023, shahzadLip2022, haliassosLips2021, mittalEmotions2020, qianThinking2020}.
Over the past few years, many datasets (e.g., ~\cite{korshunovDeepFakes2018, heForgeryNet2021, narayanDFPlatter2023}) have been proposed to support the research on deepfake detection.
A comprehensive list of the relevant publicly available datasets is given in Table~\ref{tab:datasets}.
Most of the available datasets provide examples of face manipulations through either face swapping~\cite{korshunovDeepFakes2018, dolhanskyDeepFake2020, zhouFace2021} or face reenactment~\cite{kwonKoDF2021, khalidFakeAVCeleb2021} techniques.
In terms of the number of samples, earlier datasets are smaller due to the limited availability of face manipulation techniques.
With the rapid advancements in content generation technology, several large-scale datasets such as DFDC~\cite{dolhanskyDeepFake2020}, DeeperForensics~\cite{jiangDeeperForensics12020}, KoDF~\cite{kwonKoDF2021}, and DF-Platter~\cite{narayanDFPlatter2023} have been proposed.
However, the task associated with these datasets is mainly restricted to coarse-level deepfake detection.
Until this point manipulations are mainly applied only to the visual modality, and later, audio manipulations~\cite{liuASVspoof2023} and audio-visual manipulations~\cite{khalidFakeAVCeleb2021} have been proposed to increase the complexity of the task.

In 2022, LAV-DF~\cite{caiYou2022} was introduced to become the first content-driven deepfake dataset for temporal localization.
However, the quality and scale of LAV-DF are limited, and the state-of-the-art methods designed for temporal localization~\cite{caiGlitch2023, zhangUMMAFormer2023} are already achieving very strong performance.
\datasetabbr{} addresses these gaps by improving the quality, diversity, and scale of the previous datasets designed for temporal deepfake localization.
Given that LAV-DF is the only available public dataset that has been designed for the same task as the dataset proposed in this paper, next we do a direct comparison of the two datasets.
In addition to the fact that \datasetabbr{} is significantly larger than LAV-DF, in terms of the number of subjects, and amount of real and fake videos, the following differences further highlight our contributions.

\begin{itemize}
\item{LAV-DF uses a rule-based system to find antonyms that maximize the change in sentiment in the transcript manipulation step.
We argue that naively choosing the antonyms causes context inconsistencies and low diversity of the fake content.
\datasetabbr{} addresses this issue with the use of a large language model, which results in diverse and context-consistent fake content.}
\item{The output quality of the visual generator Wav2Lip~\cite{prajwalLip2020} and audio generator SV2TTS~\cite{jiaTransfer2018} used for generating LAV-DF is not sufficient for state-of-the-art detection methods.
\datasetabbr{} utilizes the latest open-source state-of-the-art methods for high-quality audio and video generation.}
\item{LAV-DF includes only \emph{replacement} as a manipulation strategy.
\datasetabbr{} includes two additional challenging manipulation strategies, \textit{deletion} and \emph{insertion}.}
\end{itemize}

\section{\dataset{} Dataset}
\label{sec:Dataset}
\begin{figure*}[t]
\centering
\includegraphics[width=0.8\textwidth]{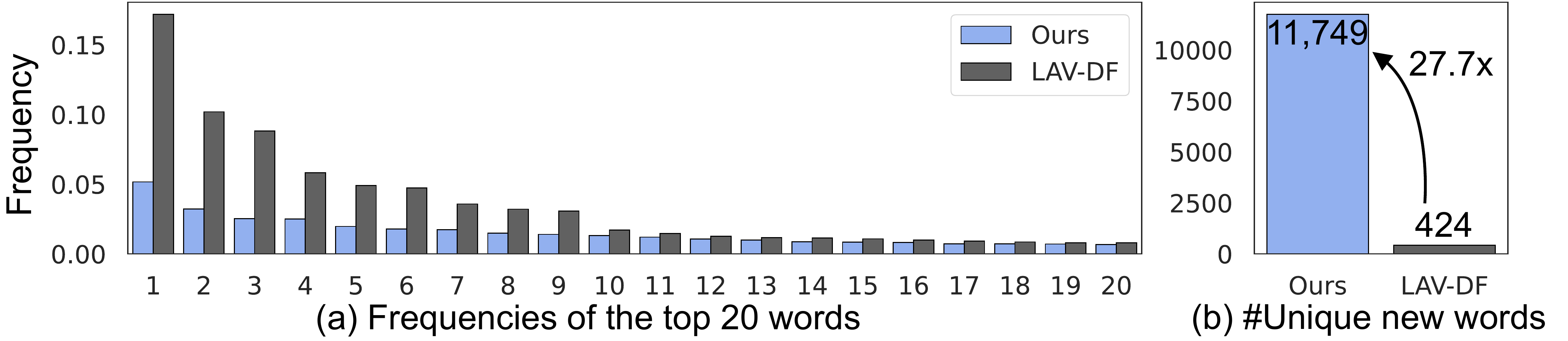}
\vspace{-2mm}
\caption{\textbf{Comparison of transcript modifications in \datasetabbr{} and LAV-DF.}}
\label{fig:word_distribution}
\end{figure*}

\datasetabbr{} is a large-scale audio-visual deepfake dataset, including 1,886 hours of audio-visual data from 2,068 unique subjects captured in diverse background environments.
This positions the proposed dataset as the most comprehensive audio-visual benchmark as illustrated in Figure~\ref{fig:teaser} and Table~\ref{tab:datasets}.
The generated videos in \datasetabbr{} preserve the background and identity present in the real videos, while the content is carefully manipulated with content-driven audio-visual data.
Following previous deepfake dataset generation research~\cite{khalidFakeAVCeleb2021, caiYou2022}, the dataset includes three different combinations of modified modalities in the generated fake videos. Please note that here we also introduce the concept of content-driven modifications for unimodal as well as multimodal aspects. We further elaborate on this in the supplementary material.

\begin{itemize}
\item{\textbf{Fake Audio} and \textbf{Fake Visual.} Both the real audio and visual frames are manipulated.}
\item{\textbf{Fake Audio} and \textbf{Real Visual.} Only the real audio corresponding to \emph{replacements} and \emph{deletions} is manipulated. To further increase data quality, the fake audio, and the corresponding length-normalized real visual segments are synchronized. As for the \emph{insertions}, new visual segments are generated based on the length of the fake audio and are lip-synced to the background noise (i.e., closed mouth).}
\item{\textbf{Real Audio} and \textbf{Fake Visual.} Only the real visual frames corresponding to \emph{replacements} and \emph{deletions} are manipulated. To further increase data quality, the length of the fake visual segments is normalized to match the length of the real audio. As for the \emph{insertions}, background noise is inserted for the corresponding fake visual segments.}
\end{itemize}


\subsection{Data Generation Pipeline}
The three-stage pipeline for generating content-driven deepfakes is illustrated in Figure~\ref{fig:pipeline}.
A subset of real videos from the Voxceleb2~\cite{chungVoxCeleb22018} dataset is pre-processed to extract the audio using FFmpeg~\cite{tomarConverting2006}, followed by Whisper-based~\cite{radfordRobust2023} real transcript generation.

\subsubsection{Transcript Manipulation}
\label{sec:transcript_manipulation} \hfill\\
\noindent \textbf{Manipulation Strategy.}
The first stage for generating content-driven deepfakes is transcript manipulation.
We utilize ChatGPT for altering the real transcripts.
Through LangChain~\cite{chaseLangChain2022} the output of ChatGPT is a structured JSON with four fields: 1) \texttt{operation:} This set contains \emph{replace}, \emph{delete}, and \emph{insert}, which has been applied on the input; 2) \texttt{old\_word:} The word in the input to \emph{replace} or \emph{delete}; 3) \texttt{new\_word:} The word in the input to \emph{insert} or \emph{replace}; 4) \texttt{index:} The location of the operation in the input.
The number of transcript modifications depends on the video length and is determined by the following equation $M = \mathbf{ceil}(t/10)$ where $M$ is the number of modifications and $t$ (sec) is the length of the video.
We followed~\cite{brownLanguage2020} and built a few-shot prompt template for ChatGPT.

\begin{prompt}[title={Prompt \thetcbcounter: Transcripts manipulation}]
\textbf{System:} You are a helpful text modifier. Your target is to modify the provided text to invert its meaning to the opposite direction. Here is the transcript of the audio. Please use the provided operations to modify the transcript to change its meaning. The operation can be one of ``delete'', ``insert'' and ``replace''.

\textbf{Human:} \{EXAMPLE INPUT 1\}

\textbf{AI:} \{EXAMPLE OUTPUT 1\}

\textbf{Human:} \{EXAMPLE INPUT 2\}

\textbf{AI:} \{EXAMPLE OUTPUT 2\}

......  

\textbf{Human:} Please generate output for the following input with \{NUM\} operations. \{INPUT\}
\end{prompt}

\noindent \textbf{Analysis.}
Figure~\ref{fig:word_distribution} (a) illustrates a comparison 
of the frequencies of the top 20 words in \datasetabbr{} and LAV-DF~\cite{caiYou2022}.
The results show that few words in LAV-DF have dominant frequencies ($>$10$\%$), whereas this issue is drastically reduced in \datasetabbr{}.
Owing to the contribution of ChatGPT, we also observed a significant increase in unique new words ($>$27.7$\times$) in the modified transcripts compared to LAV-DF, Figure~\ref{fig:word_distribution} (b).
This statistical comparison shows that the proposed LLM-based transcript manipulation strategy generates more diverse content compared to the rule-based strategy employed in LAV-DF.
We further elaborate on the advantages of using an LLM in this step in the supplementary material.

\begin{figure*}[t]
\centering
\includegraphics[width=\textwidth]{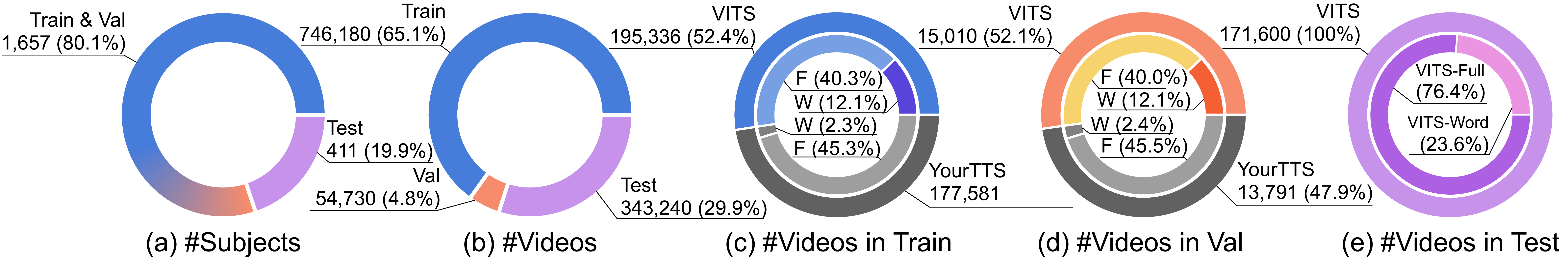}\vspace{-2mm}
\caption{\textbf{Data partitioning in \datasetabbr{}.} \textmd{(a) The number of subjects in the \emph{train}, \emph{validation}, and \emph{test} sets. (b) The number of videos in the \emph{train}, \emph{validation}, and \emph{test} sets. (c) The number of videos with different audio generation methods in the \emph{train} set. (d) The number of videos with different audio generation methods in the \emph{validation} set. (e) The number of videos with different audio generation methods in the \emph{test} set. F denotes audio generation for the \emph{full} transcript and cropping of the \texttt{new\_word}(s) and W denotes audio generation only for the \texttt{new\_word}(s).}}
\label{fig:dataset_stats}
\end{figure*}

\subsubsection{Audio Generation}
\label{sec:audio_generation} \hfill\\
\noindent \textbf{Manipulation Strategy.}
The next stage is to generate high-quality audio with the same style as the speaker.
The audio is first separated into background noise and speech using Denoiser~\cite{defossezReal2020}.
Zero-shot voice cloning methods such as SV2TTS~\cite{jiaTransfer2018} utilized by previous datasets~\cite{khalidFakeAVCeleb2021, caiYou2022} have low signal-to-noise ratio resulting in low-quality audio manipulations that are easily localized by BA-TFD~\cite{caiGlitch2023} and UMMAFormer~\cite{zhangUMMAFormer2023}.
To increase the quality of the generated audio, we employ the identity-dependent text-to-speech method VITS~\cite{kimConditional2021} for a subset of the subjects.
Further diversity in the audio generation was introduced by utilizing the identity-independent text-to-speech method YourTTS~\cite{casanovaYourTTS2022} for the rest of the subjects.

Audio generation is slightly different for each of the manipulation strategies (i.e., \emph{replace}, \emph{insert} and \emph{delete}).
In the case of \emph{replace} and \emph{insert}, we need to generate new audio corresponding to \texttt{new\_word}(s).
Generally, there are two ways to generate the \texttt{new\_word}(s):
1) Generate audio for the final fake transcript and crop it to get the audio for the required \texttt{new\_word}(s) and 2) Generate audio only for the \texttt{new\_word}(s).
To bring further diversity and challenge, we use both strategies to generate audio for the \texttt{new\_word}(s).
In the case of \emph{delete}, only the background noise is retained.
After the audio manipulation, we normalized the loudness of the fake audio segments to the original audio to add more realism.
To keep the consistency with the environmental noise, we add the background noise previously separated to the final audio output.

\begin{table}[t]
\centering
\caption{\textbf{Audio quality of \datasetabbr{}.} \textmd{Quality of the generated audio in terms of SECS, SNR and FAD.}}\vspace{-2mm}
\scalebox{0.85}{
\begin{tabular}{l|ccc}
\toprule[0.4mm]
\rowcolor{mygray} \textbf{Dataset} & \textbf{SECS($\uparrow$)} & \textbf{SNR($\uparrow$)} & \textbf{FAD($\downarrow$)} \\ \hline \hline
FakeAVCeleb~\cite{khalidFakeAVCeleb2021} & 0.543 & 2.16 & 6.598 \\
LAV-DF~\cite{caiYou2022} & 0.984 & 7.83 & 0.306 \\ \hline
\datasetabbr{} (Train) & 0.991 & 9.40 & 0.091 \\
\datasetabbr{} (Validation) & 0.991 & 9.16 & 0.091 \\
\datasetabbr{} (Test) & 0.991 & 9.42 & 0.083  \\
\datasetabbr{} (Overall) & \textbf{0.991} & \textbf{9.39} & \textbf{0.088} \\
\bottomrule[0.4mm]
\end{tabular}}
\label{tab:audio_quality}
\end{table}

\noindent \textbf{Analysis.}
We evaluated the quality of the audio generation following previous works~\cite{choiAttentron2020, casanovaSCGlowTTS2021} (note that for all datasets, we only evaluated the samples where the audio modality is modified).
The results are shown in Table~\ref{tab:audio_quality}.
The first evaluation metric is speaker encoder cosine similarity (SECS)~\cite{wanGeneralized2018}.
It measures the similarity of the speakers given a pair of audio in the range $[-1, 1]$.
We also calculated the signal-to-noise ratio (SNR) for all fake audio and Fr\'echet audio distance (FAD)~\cite{kilgourFrechet2019}.
The results indicate that \datasetabbr{} contains higher quality audio compared to other datasets.

\begin{figure*}[t]
\centering
\includegraphics[width=\textwidth]{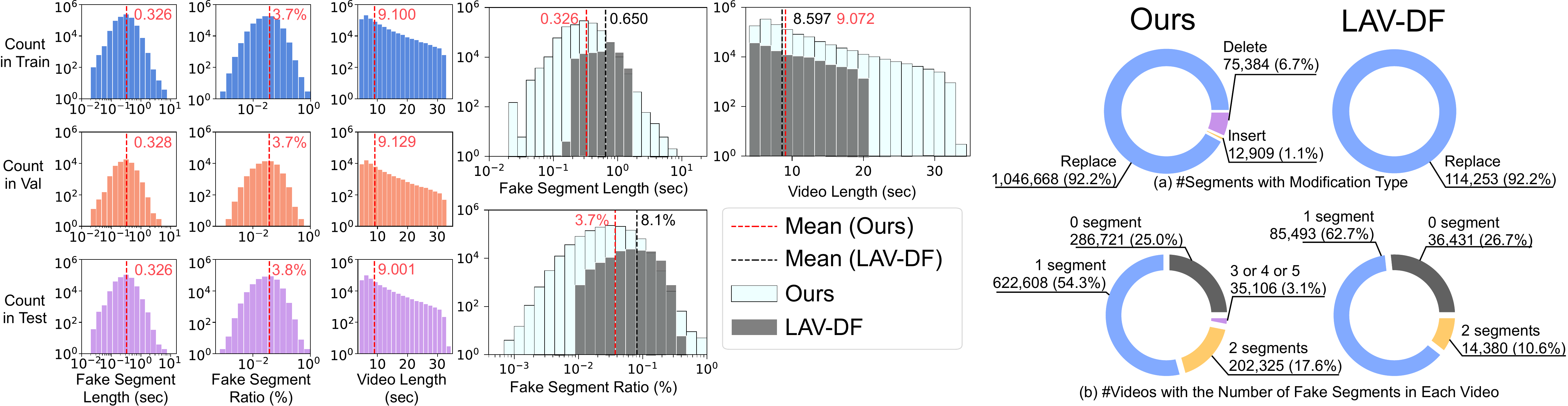}\vspace{-2mm}
\caption{\textbf{Comparison of \datasetabbr{} and LAV-DF.} \textmd{The left three-row three-column histograms illustrate the fake segment absolute lengths (sec), the fake segment lengths proportion in videos (\%) and the video lengths (sec) in the \emph{train}, \emph{validation}, and \emph{test} sets. In the middle, the histograms illustrate the overall statistics for fake segment lengths, proportions and video lengths, compared with LAV-DF. For the fake segment lengths and proportions, the X-axis is in log scale and for video lengths, the X-axis is in linear scale. For all histograms, the Y-axis is in linear scale. The vertical dotted lines and numbers in histograms represent the mean value. On the right side, (a) The number of segments with different modifications and (b) The number of videos with different numbers of segments per video.}}
\label{fig:segment_length_comparison}
\end{figure*}

\subsubsection{Video Generation} \hfill\\
\noindent \textbf{Manipulation Strategy.}
The final stage of the generation pipeline is visual content generation.
After the audio is generated, the lip-synced visual frames are generated based on the subjects' original pose and the fake audio.
We investigated several face reenactment strategies including EAMM~\cite{jiEAMM2022}, AVFR-GAN~\cite{agarwalAudioVisual2023}, DiffTalk~\cite{shenDiffTalk2023}, AD-NeRF~\cite{guoADNeRF2021} and ATVGnet~\cite{chenHierarchical2019} and concluded that these methods are not well suited for zero-shot lip-synced generation of unseen speakers.
Thus, we use TalkLip~\cite{wangSeeing2023} for visual content generation which is primarily designed for zero-shot lip-sync scenarios.
LipTalk is 1) Identity-independent, 2) Lip-syncing only without generating new poses, 3) Fast, 4) State-of-the-art, and 5) Open-source.
This way we avoid the weaknesses of the aforementioned face reenactment strategies.
The pre-trained TalkLip model is used to generate fake visual frames that are lip-synchronized with the input audio and can be used for \emph{insertion}, \emph{replacement}, and \emph{deletion}.

\begin{table}[t]
\centering
\caption{\textbf{Visual quality of \datasetabbr{}.} \textmd{Quality of the generated video in terms of PSNR, SSIM and FID.}}\vspace{-2mm}
\scalebox{0.85}{
\begin{tabular}{l|ccc}
\toprule[0.4mm]
\rowcolor{mygray} \textbf{Dataset} & \textbf{PSNR($\uparrow$)} & \textbf{SSIM($\uparrow$)} & \textbf{FID($\downarrow$)} \\ \hline \hline
FF++~\cite{rosslerFaceForensics2019} & 24.40 & 0.812 & 1.06 \\ 
DFDC~\cite{dolhanskyDeepFake2020} & - & - & 5.69 \\ 
FakeAVCeleb~\cite{khalidFakeAVCeleb2021} & 29.82 & 0.919 & 2.29 \\
LAV-DF~\cite{caiYou2022} & 33.06 & 0.898 & 1.92 \\ \hline
\datasetabbr{} (Train) &  39.50  &  0.977  &  0.50  \\
\datasetabbr{} (Validation) &  39.54  &  0.977  &  0.49  \\
\datasetabbr{} (Test) &  39.48  &  0.977  &  0.56  \\
\datasetabbr{} (Overall) &  \textbf{39.49}  &  \textbf{0.977}  & \textbf{0.49} \\
\bottomrule[0.4mm]
\end{tabular}}
\label{tab:visual_quality}
\vspace{-4mm}
\end{table}

\noindent \textbf{Analysis.}
To evaluate the visual quality of \datasetabbr{}, we used peak signal-to-noise ratio (PSNR), structural similarity index (SSIM)~\cite{wangImage2004} and Fr\'echet inception distance (FID)~\cite{heuselGANs2017} metrics as shown in Table~\ref{tab:visual_quality}.
Note that for a fair comparison, we pre-processed the videos to a common format.
The videos of FF++~\cite{rosslerFaceForensics2019} and DFDC~\cite{dolhanskyDeepFake2020} are `in-the-wild', whereas FakeAVCeleb~\cite{khalidFakeAVCeleb2021}, LAV-DF~\cite{caiYou2022} and \datasetabbr{} are facial videos.
Thus, we cropped the facial region for FF++ and DFDC for visual quality assessment.
Since FakeAVCeleb, LAV-DF and \datasetabbr{} are multimodal, for a fair comparison, we only used samples with the visual modality modified.
The results indicate that \datasetabbr{} is of higher visual quality compared to existing datasets.

\subsection{Dataset Statistics}
\label{sec:statistics}
We split the dataset into \textit{train}, \textit{validation}, and \textit{test} sets.
We first randomly select 1,657 subjects for the \textit{train} set and 411 subjects for the \textit{test} set without overlap.
The \textit{validation} set is randomly selected from the \textit{train} set.
The \emph{test} set contains only samples with VITS-based identity-dependent audio.
The variation in the number of subjects and videos in the sets is presented in Table~\ref{tab:dataset_stats} and Figure~\ref{fig:dataset_stats}.

Figure~\ref{fig:segment_length_comparison} illustrates the direct comparison of \datasetabbr{} and LAV-DF.
The results indicate that \datasetabbr{} is more diverse in terms of modifications, subjects, fake segment and video lengths, and a lower average proportion of fake segments, making the dataset a vital asset for building better deepfake localization methods.

\begin{table}[t]
\centering
\caption{\textbf{Number of subjects and videos in \datasetabbr{}.}}
\vspace{-2mm}
\scalebox{0.85}{
\begin{tabular}{l|cccc}
\toprule[0.4mm]
\rowcolor{mygray} \textbf{Subset} & \textbf{\#Subjects} & \textbf{\#Real Videos} & \textbf{\#Fake Videos} & \textbf{\#Videos} \\ \hline \hline
Train & \multirow{2}{*}{1,657} & 186,666 & 559,514 & 746,180  \\
Validation & & 14,235 & 43,105 & 54,730 \\
Test & 411 & 85,820 & 257,420 & 343,240 \\ \hline
Overall & 2,068 & 286,721 & 860,039 & 1,146,760 \\
\bottomrule[0.4mm]
\end{tabular}}
\label{tab:dataset_stats}
\end{table}

\subsection{Human Quality Assessment}
\label{sec:human_quality_assessment}
To investigate if humans can detect the deepfakes in \datasetabbr{}, we also conducted a user study with 25 participants with prior experience in video manipulation in the computer vision domain (note that the authors did not participate in the study)\footnote{All procedures in this study were conducted in accordance with Monash University Human Research Ethics Committee approval 41545.}.
200 random samples that contain 0 or 1 modification were selected for the study, where 100 from LAV-DF and 100 from \datasetabbr{}.
Each participant was asked to classify 20 videos (5 real and 5 fake from LAV-DF dataset, 5 real and 5 fake from \datasetabbr{}) as real or fake and propose the potential fake segment start and end point.
The user study results presented in Table~\ref{tab:user_study} indicate that the deepfake content in \datasetabbr{} is very challenging to detect for humans, and \datasetabbr{} is more difficult than LAV-DF.

\subsection{Computational Cost}
We spent around $\sim$600 GPU hours for speech recognition with Whisper~\cite{radfordRobust2023}, $\sim$2100 GPU hours for training VITS~\cite{kimConditional2021} (each of the 721 VITS models requires $\sim$3hrs), and $\sim$300 GPU hours for data generation.
Overall, we needed $\sim$3000 GPU hours to generate \datasetabbr{} with NVIDIA RTX6000 GPUs.

\section{Benchmarks and Metrics}
\label{sec:benchmark}

This section outlines the benchmark protocol for \datasetabbr{} along with the used evaluation metrics.
The goal is to detect and localize content-driven audio, visual, and audio-visual manipulations.

\begin{table}[t]
\centering
\caption{\textbf{User study results for \datasetabbr{} and LAV-DF.}}
\vspace{-2mm}
\scalebox{0.85}{
\begin{tabular}{l|cccc}
\toprule[0.4mm]
\rowcolor{mygray} \textbf{User Study} & \textbf{Acc.} & \textbf{AP@0.1} & \textbf{AP@0.5} & \textbf{AR@1} \\ \hline \hline
LAV-DF & 84.03 & 36.80 & 14.17 & 10.04 \\
\datasetabbr{} & 68.64 & 15.32 & 01.92 & 02.54 \\
\bottomrule[0.4mm]
\end{tabular}}
\label{tab:user_study}
\vspace{-3mm}
\end{table}

\subsection{Data Partitioning}
\label{sec:data_partitioning}
The dataset is organized in \emph{train}, \emph{validation}, and \emph{test} sets, as described in Section~\ref{sec:statistics}.
The original \emph{test} set (all modifications) is referred to as \emph{fullset} in the rest of the text.
For a fair comparison with visual-only and audio-only methods, we also prepared \emph{subset V} (by excluding the videos with audio-only modifications from \emph{fullset}) and \emph{subset A} (by excluding the videos with visual-only modifications from \emph{fullset}).

\begin{table*}[t]
\centering
\caption{\textbf{Temporal deepfake localization benchmark.} \textmd{Performance comparison of state-of-the-art methods on the proposed \datasetabbr{} dataset. The results are significantly low, indicating that \datasetabbr{} is an important benchmark for this task.}}
\scalebox{0.8}{
\begin{tabular}{c|l|c|cccc|ccccc}
\toprule[0.4mm]
\rowcolor{mygray} \textbf{Set} &\textbf{Method} & \textbf{Mod.} & \textbf{AP@0.5} & \textbf{AP@0.75} & \textbf{AP@0.9}	& \textbf{AP@0.95}& \textbf{AR@50} & \textbf{AR@30}& \textbf{AR@20}& \textbf{AR@10}& \textbf{AR@5}\\ \hline \hline
\multirow{11}{*}{\rotatebox[origin=c]{90}{\textbf{Fullset}}} & PyAnnote (Zero-Shot)~\cite{plaquetPowerset2023} & A  & 00.03 & 00.00 & 00.00 & 00.00 & 00.67 & 00.67 & 00.67 & 00.67 & 00.67 \\
& Meso4~\cite{afcharMesoNet2018} & V & 09.86 & 06.05 & 02.22 & 00.59 & 38.92 & 38.91 & 38.81 & 36.47 & 26.91 \\
&MesoInception4~\cite{afcharMesoNet2018} & V & 08.50 & 05.16 & 01.89 & 00.50 & 39.27 & 39.22 & 39.00 & 35.78 & 24.59 \\
&EfficientViT~\cite{coccominiCombining2022} & V & 14.71 & 02.42 & 00.13 & 00.01 & 27.04 & 26.99 & 26.43 & 23.90 & 20.31 \\
&TriDet + VideoMAEv2~\cite{shiTriDet2023, wangVideoMAE2023} & V  & 21.67 & 05.83 & 00.54 & 00.06 & 20.27 & 20.23 & 20.12 & 19.50 & 18.18 \\
&TriDet + InternVideo~\cite{shiTriDet2023, wangInternVideo2022} & V  & 29.66 & 09.02 & 00.79 & 00.09 & 24.08 & 24.06 & 23.96 & 23.50 & 22.55 \\
&ActionFormer + VideoMAEv2~\cite{zhangActionFormer2022, wangVideoMAE2023} & V  & 20.24 & 05.73 & 00.57 & 00.07 & 19.97 & 19.93 & 19.81 & 19.11 & 17.80 \\
&ActionFormer + InternVideo~\cite{zhangActionFormer2022, wangInternVideo2022} & V  & 36.08 & 12.01 & 01.23 & 00.16 & 27.11 & 27.08 & 27.00 & 26.60 & 25.80 \\
&BA-TFD~\cite{caiYou2022} & AV & 37.37 & 06.34 & 00.19 & 00.02 & 45.55 & 40.37 & 35.95 & 30.66 & 26.82 \\
&BA-TFD+~\cite{caiGlitch2023} & AV & 44.42 & 13.64 & 00.48 & 00.03 & \textbf{48.86} & \textbf{44.51} & 40.37 & 34.67 & 29.88 \\
&UMMAFormer~\cite{zhangUMMAFormer2023} & AV & \textbf{51.64} & \textbf{28.07} & \textbf{07.65} & \textbf{01.58} & 44.07 & 43.93 & \textbf{43.45} & \textbf{42.09} & \textbf{40.27} \\ \hline \hline
\multirow{11}{*}{\rotatebox[origin=c]{90}{\textbf{Subset V}}} & PyAnnote (Zero-Shot)~\cite{plaquetPowerset2023} & A & 00.02 & 00.00 & 00.00 & 00.00 & 00.52 & 00.52 & 00.52 & 00.52 & 00.52 \\
&Meso4~\cite{afcharMesoNet2018} & V & 15.31 & 09.54 & 03.52 & 00.93 & 58.04 & 58.03 & 57.87 & 54.37 & 40.06 \\
&MesoInception4~\cite{afcharMesoNet2018} & V & 13.38 & 08.25 & 03.05 & 00.81 & 58.54 & 58.48 & 58.15 & 53.34 & 36.59 \\
&EfficientViT~\cite{coccominiCombining2022} & V & 23.21 & 03.92 & 00.21 & 00.02 & 37.52 & 37.46 & 36.88 & 34.19 & 29.64 \\
&TriDet + VideoMAEv2~\cite{shiTriDet2023, wangVideoMAE2023} & V & 26.45 & 07.35 & 00.74 & 00.08 & 22.49 & 22.47 & 22.42 & 22.04 & 21.09 \\
&TriDet + InternVideo~\cite{shiTriDet2023, wangInternVideo2022} & V & 37.90 & 12.15  & 01.12 & 00.13  & 28.08 & 28.07 & 28.03 & 27.79 & 27.17 \\
&ActionFormer + VideoMAEv2~\cite{zhangActionFormer2022, wangVideoMAE2023} & V & 24.80 & 07.25 & 00.77 & 00.09 & 22.26 & 22.23 & 22.16 & 21.70 & 20.71 \\
&ActionFormer + InternVideo~\cite{zhangActionFormer2022, wangInternVideo2022} & V & 45.57 & 16.07 & 01.75 & 00.23 & 31.78 & 31.77 & 31.73 & 31.56 & 31.14 \\
&BA-TFD~\cite{caiYou2022} & AV & 55.34 & 09.48 & 00.30 & 00.03 & 62.66 & 55.48 & 49.53 & 43.15 & 38.48 \\
&BA-TFD+~\cite{caiGlitch2023} & AV & \textbf{65.85} & 20.37 & 00.73 & 00.05 & \textbf{65.13} & \textbf{59.07} & \textbf{53.57} & \textbf{46.79} & \textbf{41.69} \\
&UMMAFormer~\cite{zhangUMMAFormer2023} & AV & 39.07 & \textbf{20.77} & \textbf{05.62} & \textbf{01.16} & 40.39 & 40.19 & 39.51 & 37.53 & 34.93 \\ \hline \hline
\multirow{11}{*}{\rotatebox[origin=c]{90}{\textbf{Subset A}}} & PyAnnote (Zero-Shot)~\cite{plaquetPowerset2023} & A  & 00.05 & 00.01 & 00.00 & 00.00 & 00.97 & 00.97 & 00.97 & 00.97 & 00.96 \\
&Meso4~\cite{afcharMesoNet2018} & V & 07.13 & 04.17 & 01.45 & 00.39 & 29.34 & 29.34 & 29.27 & 27.58 & 20.54 \\
&MesoInception4~\cite{afcharMesoNet2018} & V & 05.88 & 03.46 & 01.19 & 00.32 & 29.46 & 29.42 & 29.26 & 26.95 & 18.80 \\
&EfficientViT~\cite{coccominiCombining2022} & V & 09.91 & 15.79 & 00.08 & 00.01 & 21.47 & 21.42 & 20.87 & 18.43 & 15.39 \\
&TriDet + VideoMAEv2~\cite{shiTriDet2023, wangVideoMAE2023} & V & 17.45 & 04.01 & 00.24 & 00.02 & 18.47 & 18.43 & 18.28 & 17.53 & 16.02 \\
&TriDet + InternVideo~\cite{shiTriDet2023, wangInternVideo2022} & V & 24.95 & 06.85 & 00.47 & 00.05 & 21.79 & 21.76 & 21.64 & 21.07 & 19.95 \\
&ActionFormer + VideoMAEv2~\cite{zhangActionFormer2022, wangVideoMAE2023} & V & 16.22 & 03.95 & 00.28 & 00.03 & 18.11 & 18.07 & 17.92 & 17.10 & 15.59 \\
&ActionFormer + InternVideo\cite{zhangActionFormer2022, wangInternVideo2022} & V & 30.86 & 09.47 & 00.78 & 00.09 & 24.49 & 24.46 & 24.36 & 23.85 & 22.87 \\
&BA-TFD~\cite{caiYou2022} & AV & 27.79 & 04.31 & 00.12 & 00.01 & 36.71 & 32.50 & 28.82 & 24.02 & 20.58 \\
&BA-TFD+~\cite{caiGlitch2023} & AV & 33.23 & 10.07 & 00.36 & 00.03 & 40.54 & 37.07 & 33.63 & 28.50 & 23.82 \\
&UMMAFormer~\cite{zhangUMMAFormer2023} & AV & \textbf{68.68} & \textbf{40.00} & \textbf{11.32} & \textbf{02.35} & \textbf{51.44} & \textbf{51.41} & \textbf{51.35} & \textbf{51.23} & \textbf{50.95} \\
\bottomrule[0.4mm]
\end{tabular}
}
\label{tab:temporal_localization}
\end{table*}

\subsection{Implementation Details}
\label{sec:implementation_details}
For benchmarking temporal deepfake localization, we consider the following state-of-the-art methods: Pyannote~\cite{plaquetPowerset2023} is a pre-trained speaker diarization method.
TriDet~\cite{shiTriDet2023} and ActionFormer~\cite{zhangActionFormer2022} are the state-of-the-art in the temporal action localization domain.
Since these two methods require pre-trained features, we extracted the state-of-the-art features VideoMAEv2~\cite{wangVideoMAE2023} and InternVideo~\cite{wangInternVideo2022} for benchmarking.
BA-TFD~\cite{caiYou2022}, BA-TFD+~\cite{caiGlitch2023}, and UMMAFormer~\cite{zhangUMMAFormer2023} are the state-of-the-art methods specifically designed for audio-visual temporal deepfake localization.
We followed the original settings for BA-TFD and BA-TFD+.
For UMMAFormer~\cite{zhangUMMAFormer2023}, we implemented it using the InternVideo~\cite{wangInternVideo2022} visual features and BYOL-A~\cite{niizumiBYOL2021} audio features.
For image-based classification methods, we consider Meso4~\cite{afcharMesoNet2018}, MesoInception4~\cite{afcharMesoNet2018}, Xception~\cite{cholletXception2017}, Face X-Ray~\cite{liFace2020}, LipForensics~\cite{haliassosLips2021},
EfficientViT~\cite{coccominiCombining2022}, and SBI~\cite{shioharaDetecting2022}. 
We followed the procedure used in previous works~\cite{zhaoTemporal2017, caiGlitch2023} to aggregate the frame-level predictions to segments for localization.

For benchmarking deepfake detection, we trained the image-based models Meso4~\cite{afcharMesoNet2018}, MesoInception4~\cite{afcharMesoNet2018}, Xception~\cite{cholletXception2017} and EfficientViT~\cite{coccominiCombining2022} with video frames along with the corresponding labels.
For the segment-based methods MDS~\cite{chughNot2020} and MARLIN~\cite{caiMARLIN2023}, we used a sliding window to sample segments from the video for training and inference.
During the inference stage, the frame- and segment-level predictions are aggregated to video-level by \emph{max} voting.
The aggregation strategy is discussed in Section~\ref{sec:result}.
We also evaluated the zero-shot performance of several methods, including the LLM-based Video-LLaMA~\cite{zhangVideoLLaMA2023}, audio pre-trained CLAP~\cite{wuLargeScale2023}, M2TR~\cite{wangM2TR2022} and LipForensics~\cite{haliassosLips2021} pre-trained on FF++~\cite{rosslerFaceForensics2019}, Face X-Ray~\cite{liFace2020} and SBI~\cite{shioharaDetecting2022} pretrained on blending images.
For Video-LLaMA, we also evaluated 5 model ensembles (the majority vote of 5 model inferences). To investigate the impact of the level of label access, we designed 3 different label access levels for training: \emph{frame-level} labels, \emph{segment-level} labels only, and \emph{video-level} labels only.

\subsection{Evaluation Metrics}
\label{sec:evaluation_metrics}
\noindent \textbf{Temporal Deepfake Localization.}
We use average precision (AP) and average recall (AR) as prior works~\cite{heForgeryNet2021, caiYou2022}.

\noindent \textbf{Deepfake Detection.} We use standard evaluation protocol~\cite{rosslerFaceForensics2019, dolhanskyDeepFake2020} to report video-level accuracy (Acc.) and area under the curve (AUC).

\section{Results and Analysis}
\label{sec:result}

This section reports the performance of the state-of-the-art deepfake detection and localization methods described in Section~\ref{sec:implementation_details} on \datasetabbr{}.
The reported performance is based on different subsets, described in Section~\ref{sec:data_partitioning}, and different levels of label access during training, described in Section~\ref{sec:implementation_details}.

\subsection{Audio-Visual Temporal Deepfake Localization}
The results of this benchmark are depicted in Table~\ref{tab:temporal_localization}.
All state-of-the-art methods achieve significantly lower performance compared to the performance reported on previous datasets~\cite{heForgeryNet2021, caiYou2022}.
This significant drop indicates that existing temporal deepfake localization methods are falling behind with the rapid advancements in content generation.
In other words, we can claim that the highly realistic fake content in \datasetabbr{} will open an avenue for further research on temporal deepfake localization methods.

\subsection{Audio-Visual Deepfake Detection}
\label{sec:audio-visual_deepfake_detection}
Similarly to temporal deepfake localization, the results of the classical deepfake detection benchmark are summarized in Table~\ref{tab:classification}.
Models that have access only to the video-level labels during training and the zero-shot models all perform poorly on this task, except the Face X-Ray and SBI which are designed to be generalizable.
Providing the fine-grained segment-level and frame-level labels during training brings an improvement in performance.
However, even with the frame-level labels provided during training, the AUC of the best-performing methods is less than 70, due to the multimodal modifications present in \datasetabbr{}.

\begin{table*}[t]
\centering
\caption{\textbf{Deepfake detection benchmark.} \textmd{Performance comparison of state-of-the-art methods on the proposed \datasetabbr{} dataset using different evaluation protocols. E5: Ensemble 5.}}
\scalebox{0.8}{
\begin{tabular}{l|l|c|cc|cc|cc}
\toprule[0.4mm]
\rowcolor{mygray} \textbf{Label Access} & \textbf{Methods}  & \textbf{Mod.} & \multicolumn{2}{c|}{\textbf{Fullset}} & \multicolumn{2}{c|}{\textbf{Subset V}}  & \multicolumn{2}{c}{\textbf{Subset A}} \\ 
\rowcolor{mygray} \textbf{For Training}  & &  & AUC & Acc.  & AUC & Acc.  & AUC & Acc. \\ \hline \hline
Zero-Shot & Video-LLaMA (7B)~\cite{zhangVideoLLaMA2023} & AV & 50.09 & 25.23 & 50.13 & 33.51 & 50.08 & 33.49 \\
 & Video-LLaMA (13B)~\cite{zhangVideoLLaMA2023} & AV & 49.50 & 25.02 & 49.53 & 33.35 & 49.30 & 33.36 \\
 & Video-LLaMA (7B) E5~\cite{zhangVideoLLaMA2023}  & AV & 49.97 & 25.32 & 50.01 & 33.57 & 49.98 & 33.62 \\
 & Video-LLaMA (13B) E5~\cite{zhangVideoLLaMA2023} & AV & 50.74 & 25.05 & 50.52 & 33.36 & 50.78 & 33.40 \\
 & CLAP~\cite{wuLargeScale2023} & A & 50.83 & 31.99 & 50.91 & 37.83 & 50.67 & 37.54 \\
 & M2TR~\cite{wangM2TR2022} & V & 50.18 & \textbf{74.99} & 50.24 & \textbf{66.67} & 50.14 & \textbf{66.66} \\ 
 & LipForensics~\cite{haliassosLips2021} & V & 51.57 & 68.84 & 54.37 & 64.13 & 50.65 & 62.19 \\ 
 & Face X-Ray~\cite{liFace2020} & V & \textbf{61.54} & 73.83 & \textbf{61.88} & 66.59 & \textbf{60.86} & 66.35 \\ 
 & SBI~\cite{shioharaDetecting2022} & V & 55.10 & 34.04 & 57.75 & 41.51 & 53.81 & 39.38 \\ 
 \hline
Video-level & Meso4~\cite{afcharMesoNet2018} & V & 50.22 & \textbf{75.00} & 50.31 & 66.66 & 50.17 & \textbf{66.66} \\
 & MesoInception4~\cite{afcharMesoNet2018} & V & 50.05 & 75.00 & 50.01 & 66.66 & 50.06 & 66.66 \\ 
 & SBI~\cite{shioharaDetecting2022} & V & \textbf{65.82} & 69.00 & \textbf{67.31} & \textbf{67.19} & \textbf{65.11} & 65.55 \\ \hline
Segment-level & Meso4~\cite{afcharMesoNet2018} & V & 54.53 & 55.83 & 56.81 & \textbf{56.78} & 53.34 & 53.89 \\
 & MesoInception4~\cite{afcharMesoNet2018} & V & 57.16 & 28.24 & \textbf{62.14} & 37.41 & 54.64 & 35.46\\ 
 & MDS~\cite{chughNot2020}   & AV & 56.57 & \textbf{59.44} & 54.21 & 53.70 & \textbf{56.92} & \textbf{58.88} \\
 & MARLIN~\cite{caiMARLIN2023}   & V & \textbf{58.03} & 29.01 & 61.57 & 38.28 & 56.23 & 35.99 \\ \hline
Frame-level & Meso4~\cite{afcharMesoNet2018} & V & 63.05 & 49.51 & 76.30 & 64.62 & 56.27 & 47.82 \\
 & MesoInception4~\cite{afcharMesoNet2018} & V & 64.04 & 54.13 & 80.67 & 69.88 & 56.28 & 51.73 \\
 & Xception~\cite{cholletXception2017} & V & \textbf{68.68} & 61.33 & \textbf{81.97} & \textbf{81.39} & \textbf{63.19} & 57.45 \\
 & EfficientViT~\cite{coccominiCombining2022}  & V & 65.51 & \textbf{71.80} & 76.74 & 70.89  & 59.75 & \textbf{63.51} \\
\bottomrule[0.4mm]
\end{tabular}
}
\label{tab:classification}
\end{table*}

\begin{table*}[t]
\centering
\caption{\textbf{Temporal localization results on the \datasetabbr{} and LAV-DF.}}
\scalebox{0.8}{
\begin{tabular}{lc||ccc|ccc}
\toprule[0.4 mm]
\rowcolor{mygray}\textbf{Method} & \textbf{Dataset} & \textbf{AP@0.5} & \textbf{AP@0.75} & \textbf{AP@0.95} & \textbf{AR@50} & \textbf{AR@20} & \textbf{AR@10} \\ \hline \hline
\multirow{2}{*}{BA-TFD~\cite{caiYou2022}} & LAV-DF~\cite{caiYou2022} & 79.15 & 38.57 & 00.24 & 64.18 & 60.89 & 58.51 \\
& \datasetabbr{} & 37.37 & 06.34 & 00.02 & 45.55 & 35.95 & 30.66 \\ \hline
\multirow{2}{*}{BA-TFD+~\cite{caiGlitch2023}} & LAV-DF~\cite{caiYou2022} & 96.30 & 84.96 & 04.44 & 80.48 & 79.40 & 78.75 \\
& \datasetabbr{} & 44.42 & 13.64 & 00.03 & 48.86 & 40.37 & 34.67 \\ \hline
\multirow{2}{*}{UMMAFormer~\cite{zhangUMMAFormer2023}} & LAV-DF~\cite{caiYou2022} & 98.83 & 95.54 & 37.61 & 92.47 & 92.42 & 92.10 \\
& \datasetabbr{} & 51.64 & 28.09 & 01.57 & 44.07 & 43.45 & 42.09 \\
\bottomrule[0.4mm]
\end{tabular}}
\label{tab:baseline_comparison_temporal_localization}
\end{table*}

\begin{table}[t]
\centering
\caption{\textbf{Aggregation strategies.} \textmd{AUC scores on \emph{fullset} for each method using different aggregation strategies.}}
\label{tab:aggregation_strategy}
\scalebox{0.8}{
\begin{tabular}{l|ccccc}
\toprule[0.4mm]
\rowcolor{mygray} \textbf{Method $\rightarrow$} & \textbf{Meso4} & \textbf{MesoInc4} & \textbf{Xception} & \textbf{EfficientViT} & \textbf{MARLIN}\\
\rowcolor{mygray} \textbf{Strategy $\downarrow$} & \cite{afcharMesoNet2018} & \cite{afcharMesoNet2018} & \cite{cholletXception2017} & \cite{coccominiCombining2022} & \cite{caiMARLIN2023} \\ \hline \hline
max & \textbf{63.05} & \textbf{64.04} & 68.68 & \textbf{65.51} & \textbf{58.03} \\
avg & 55.61 & 54.07 & 61.44 & 58.75 & 53.20 \\
avg of top5 & 62.32 & 59.82 & \textbf{68.81} & 63.60 & 56.39 \\
\bottomrule[0.4mm]
\end{tabular}}
\end{table}

The frame- and segment-based deepfake detection methods can only produce frame- and segment-level predictions.
Thus, a suitable aggregation strategy is required to generate the video-level predictions.
We investigated several popular aggregation strategies, such as \emph{max} (e.g.,~\cite{caiYou2022}), \emph{average} (e.g.,~\cite{heoDeepFake2023, wangM2TR2022, coccominiCombining2022}), and the \emph{average of the highest 5 scores} (e.g.,~\cite{liExposing2019}) for video-level predictions.
The results of the experiment are presented in Table~\ref{tab:aggregation_strategy}.
The results show that \emph{max} is the optimal aggregation strategy on \datasetabbr{} for the considered deepfake detection methods.

\subsection{Unimodal Deepfake Detection and Localization}
We also evaluated the performance on \emph{subset V} and \emph{subset A}, as described in Section~\ref{sec:data_partitioning}.
As expected, all visual-only methods consistently perform better on \emph{subset V} compared to \emph{fullset} for both temporal localization and detection.
The same holds for \emph{subset A} and audio-only methods.

\begin{table}[t]
\caption{\textbf{Performance (AUC $\uparrow$) for classification baselines on \datasetabbr{} and LAV-DF.}}
\centering
\scalebox{0.8}{
\begin{tabular}{ll||cc}
\toprule[0.4mm]
\rowcolor{mygray} \textbf{Label Access} & \textbf{Methods} & \textbf{AV-Deepfake1M} & \textbf{LAV-DF~\cite{caiYou2022}} \\ \hline \hline
Zero-shot & LipForensics~\cite{haliassosLips2021} & 51.57 & 73.34\\
 & Face X-Ray~\cite{liFace2020} & 61.54 & 69.65 \\ 
 & SBI~\cite{shioharaDetecting2022} & 55.10 & 62.84 \\ \hline
Video-level & SBI~\cite{shioharaDetecting2022} & 65.82 & 67.23 \\ \hline
Segment-level & MDS~\cite{chughNot2020} & 56.57 & 82.80 \\ \hline
Frame-level & Xception~\cite{cholletXception2017} & 68.68 & 83.58 \\
& EfficientViT~\cite{coccominiCombining2022} & 65.51 & 96.50 \\

\bottomrule[0.4mm]
\end{tabular}
}
\label{tab:baseline_comparison_classification}
\end{table}

\begin{table}[t]
\caption{\textbf{Transfer learning results.} \textmd{\underline{\small{\textit{Dataset for pretraining}}. }}}
\centering
\scalebox{0.8}{
\begin{tabular}{cc||cc}
\toprule[0.4mm]
\rowcolor{mygray} \multicolumn{2}{r||}{\textbf{Methods $\rightarrow$}} & \textbf{BA-TFD} & \textbf{Xception}\\
\rowcolor{mygray} \textbf{Train Data} & \textbf{Test Data} &\textbf{AP@0.5 $\uparrow$} & \textbf{AUC $\uparrow$} \\ \hline \hline
LAV-DF & LAV-DF & 79.15 & 83.58 \\
\underline{\textit{AV-Deepfake1M}}, LAV-DF & LAV-DF & 83.93 & 90.12 \\
\bottomrule[0.4mm]
\end{tabular}
}
\label{tab:finetune_results}
\end{table}

\subsection{Benchmark Comparison}

We conducted additional experiments (Tables~\ref{tab:baseline_comparison_temporal_localization} and \ref{tab:baseline_comparison_classification}) to compare the performance on temporal localization and classification on \datasetabbr{} and LAV-DF~\cite{caiYou2022}.

There is a significant drop in BA-TFD~\cite{caiYou2022} temporal localization performance as compared to LAV-DF (Table~\ref{tab:baseline_comparison_temporal_localization}). A similar pattern is also observed for BA-TFD+~\cite{caiGlitch2023} (AP@0.5 96.30 $\rightarrow$ 44.42) and UMMAFormer~\cite{zhangUMMAFormer2023} (AP@0.5 98.83 $\rightarrow$ 51.64). For classification (Table~\ref{tab:baseline_comparison_classification}), the performance of Xception~\cite{cholletXception2017}, LipForensics~\cite{haliassosLips2021}, Face X-Ray~\cite{liFace2020}, and SBI~\cite{shioharaDetecting2022}  also drops compared to LAV-DF. These additional results further validate that \datasetabbr{} is more challenging than LAV-DF.

We conduct the experiments using Xception and BA-TFD pretrained on \datasetabbr{} then finetune and evaluate on LAV-DF, shown in Table~\ref{tab:finetune_results}. We observe the performance improvements are significant for both temporal localization with BA-TFD and classification with Xception, when compared with models trained on LAV-DF from scratch.

\section{Conclusion}
\label{sec:conclusion}
This paper presents \datasetabbr{}, the largest audio-visual dataset for temporal deepfake localization.
The comprehensive benchmark of the dataset utilizing state-of-the-art deepfake detection and localization methods indicates a significant drop in performance compared to previous datasets, indicating that the proposed dataset is an important asset for building the next-generation of deepfake localization methods.

\noindent \textbf{Limitations.} Similarly to other deepfake datasets, \datasetabbr{} exhibits a misbalance in terms of the number of fake and real videos.

\noindent \textbf{Broader Impact.} Owing to the diversified and realistic, content-driven fake videos, \datasetabbr{} will support the development of robust audio-visual deepfake detection and localization models.

\noindent \textbf{Ethics Statement.} 
We acknowledge that \datasetabbr{} may raise ethical concerns such as the potential misuse of facial videos of celebrities, and even the data generation pipeline could have a potential negative impact.
Misuse could include the creation of deepfake videos or other forms of exploitation.
To avoid such issues, we have taken several measures such as distributing the data with a proper end-user license agreement, where we will impose certain restrictions on the usage of the data, such as the data generation technology and resulting content being restricted to research purposes only.

\bibliographystyle{ACM-Reference-Format}
\bibliography{main_cr}


\begin{thebibliography}{78}


\ifx \showCODEN    \undefined \def \showCODEN     #1{\unskip}     \fi
\ifx \showDOI      \undefined \def \showDOI       #1{#1}\fi
\ifx \showISBNx    \undefined \def \showISBNx     #1{\unskip}     \fi
\ifx \showISBNxiii \undefined \def \showISBNxiii  #1{\unskip}     \fi
\ifx \showISSN     \undefined \def \showISSN      #1{\unskip}     \fi
\ifx \showLCCN     \undefined \def \showLCCN      #1{\unskip}     \fi
\ifx \shownote     \undefined \def \shownote      #1{#1}          \fi
\ifx \showarticletitle \undefined \def \showarticletitle #1{#1}   \fi
\ifx \showURL      \undefined \def \showURL       {\relax}        \fi
\providecommand\bibfield[2]{#2}
\providecommand\bibinfo[2]{#2}
\providecommand\natexlab[1]{#1}
\providecommand\showeprint[2][]{arXiv:#2}

\bibitem[Afchar et~al\mbox{.}(2018)]%
        {afcharMesoNet2018}
\bibfield{author}{\bibinfo{person}{Darius Afchar}, \bibinfo{person}{Vincent Nozick}, \bibinfo{person}{Junichi Yamagishi}, {and} \bibinfo{person}{Isao Echizen}.} \bibinfo{year}{2018}\natexlab{}.
\newblock \showarticletitle{{MesoNet}: a {Compact} {Facial} {Video} {Forgery} {Detection} {Network}}. In \bibinfo{booktitle}{\emph{2018 {IEEE} {International} {Workshop} on {Information} {Forensics} and {Security} ({WIFS})}}. \bibinfo{pages}{1--7}.
\newblock
\newblock
\shownote{ISSN: 2157-4774}.


\bibitem[Agarwal et~al\mbox{.}(2023)]%
        {agarwalAudioVisual2023}
\bibfield{author}{\bibinfo{person}{Madhav Agarwal}, \bibinfo{person}{Rudrabha Mukhopadhyay}, \bibinfo{person}{Vinay~P. Namboodiri}, {and} \bibinfo{person}{C.~V. Jawahar}.} \bibinfo{year}{2023}\natexlab{}.
\newblock \showarticletitle{Audio-{Visual} {Face} {Reenactment}}. In \bibinfo{booktitle}{\emph{Proceedings of the {IEEE}/{CVF} {Winter} {Conference} on {Applications} of {Computer} {Vision}}}. \bibinfo{pages}{5178--5187}.
\newblock


\bibitem[Brown et~al\mbox{.}(2020)]%
        {brownLanguage2020}
\bibfield{author}{\bibinfo{person}{Tom Brown}, \bibinfo{person}{Benjamin Mann}, \bibinfo{person}{Nick Ryder}, \bibinfo{person}{Melanie Subbiah}, \bibinfo{person}{Jared~D Kaplan}, \bibinfo{person}{Prafulla Dhariwal}, \bibinfo{person}{Arvind Neelakantan}, \bibinfo{person}{Pranav Shyam}, \bibinfo{person}{Girish Sastry}, \bibinfo{person}{Amanda Askell}, \bibinfo{person}{Sandhini Agarwal}, \bibinfo{person}{Ariel Herbert-Voss}, \bibinfo{person}{Gretchen Krueger}, \bibinfo{person}{Tom Henighan}, \bibinfo{person}{Rewon Child}, \bibinfo{person}{Aditya Ramesh}, \bibinfo{person}{Daniel Ziegler}, \bibinfo{person}{Jeffrey Wu}, \bibinfo{person}{Clemens Winter}, \bibinfo{person}{Chris Hesse}, \bibinfo{person}{Mark Chen}, \bibinfo{person}{Eric Sigler}, \bibinfo{person}{Mateusz Litwin}, \bibinfo{person}{Scott Gray}, \bibinfo{person}{Benjamin Chess}, \bibinfo{person}{Jack Clark}, \bibinfo{person}{Christopher Berner}, \bibinfo{person}{Sam McCandlish}, \bibinfo{person}{Alec Radford}, \bibinfo{person}{Ilya Sutskever}, {and}
  \bibinfo{person}{Dario Amodei}.} \bibinfo{year}{2020}\natexlab{}.
\newblock \showarticletitle{Language {Models} are {Few}-{Shot} {Learners}}. In \bibinfo{booktitle}{\emph{Advances in {Neural} {Information} {Processing} {Systems}}}, Vol.~\bibinfo{volume}{33}. \bibinfo{publisher}{Curran Associates, Inc.}, \bibinfo{pages}{1877--1901}.
\newblock


\bibitem[Cai et~al\mbox{.}(2023a)]%
        {caiGlitch2023}
\bibfield{author}{\bibinfo{person}{Zhixi Cai}, \bibinfo{person}{Shreya Ghosh}, \bibinfo{person}{Abhinav Dhall}, \bibinfo{person}{Tom Gedeon}, \bibinfo{person}{Kalin Stefanov}, {and} \bibinfo{person}{Munawar Hayat}.} \bibinfo{year}{2023}\natexlab{a}.
\newblock \showarticletitle{Glitch in the matrix: {A} large scale benchmark for content driven audio–visual forgery detection and localization}.
\newblock \bibinfo{journal}{\emph{Computer Vision and Image Understanding}}  \bibinfo{volume}{236} (\bibinfo{date}{Nov.} \bibinfo{year}{2023}), \bibinfo{pages}{103818}.
\newblock
\showISSN{1077-3142}


\bibitem[Cai et~al\mbox{.}(2023b)]%
        {caiMARLIN2023}
\bibfield{author}{\bibinfo{person}{Zhixi Cai}, \bibinfo{person}{Shreya Ghosh}, \bibinfo{person}{Kalin Stefanov}, \bibinfo{person}{Abhinav Dhall}, \bibinfo{person}{Jianfei Cai}, \bibinfo{person}{Hamid Rezatofighi}, \bibinfo{person}{Reza Haffari}, {and} \bibinfo{person}{Munawar Hayat}.} \bibinfo{year}{2023}\natexlab{b}.
\newblock \showarticletitle{{MARLIN}: {Masked} {Autoencoder} for {Facial} {Video} {Representation} {LearnINg}}. In \bibinfo{booktitle}{\emph{Proceedings of the {IEEE}/{CVF} {Conference} on {Computer} {Vision} and {Pattern} {Recognition}}}. \bibinfo{publisher}{IEEE}, \bibinfo{address}{Vancouver, BC, Canada}, \bibinfo{pages}{1493--1504}.
\newblock
\showISBNx{979-8-3503-0129-8}


\bibitem[Cai et~al\mbox{.}(2022)]%
        {caiYou2022}
\bibfield{author}{\bibinfo{person}{Zhixi Cai}, \bibinfo{person}{Kalin Stefanov}, \bibinfo{person}{Abhinav Dhall}, {and} \bibinfo{person}{Munawar Hayat}.} \bibinfo{year}{2022}\natexlab{}.
\newblock \showarticletitle{Do {You} {Really} {Mean} {That}? {Content} {Driven} {Audio}-{Visual} {Deepfake} {Dataset} and {Multimodal} {Method} for {Temporal} {Forgery} {Localization}}. In \bibinfo{booktitle}{\emph{2022 {International} {Conference} on {Digital} {Image} {Computing}: {Techniques} and {Applications} ({DICTA})}}. \bibinfo{address}{Sydney, Australia}, \bibinfo{pages}{1--10}.
\newblock


\bibitem[Casanova et~al\mbox{.}(2021)]%
        {casanovaSCGlowTTS2021}
\bibfield{author}{\bibinfo{person}{Edresson Casanova}, \bibinfo{person}{Christopher Shulby}, \bibinfo{person}{Eren Gölge}, \bibinfo{person}{Nicolas~Michael Müller}, \bibinfo{person}{Frederico Santos~De Oliveira}, \bibinfo{person}{Arnaldo Candido~Jr.}, \bibinfo{person}{Anderson Da~Silva Soares}, \bibinfo{person}{Sandra~Maria Aluisio}, {and} \bibinfo{person}{Moacir~Antonelli Ponti}.} \bibinfo{year}{2021}\natexlab{}.
\newblock \showarticletitle{{SC}-{GlowTTS}: {An} {Efficient} {Zero}-{Shot} {Multi}-{Speaker} {Text}-{To}-{Speech} {Model}}. In \bibinfo{booktitle}{\emph{Interspeech 2021}}. \bibinfo{publisher}{ISCA}, \bibinfo{pages}{3645--3649}.
\newblock


\bibitem[Casanova et~al\mbox{.}(2022)]%
        {casanovaYourTTS2022}
\bibfield{author}{\bibinfo{person}{Edresson Casanova}, \bibinfo{person}{Julian Weber}, \bibinfo{person}{Christopher~D. Shulby}, \bibinfo{person}{Arnaldo~Candido Junior}, \bibinfo{person}{Eren Gölge}, {and} \bibinfo{person}{Moacir~A. Ponti}.} \bibinfo{year}{2022}\natexlab{}.
\newblock \showarticletitle{{YourTTS}: {Towards} {Zero}-{Shot} {Multi}-{Speaker} {TTS} and {Zero}-{Shot} {Voice} {Conversion} for {Everyone}}. In \bibinfo{booktitle}{\emph{Proceedings of the 39th {International} {Conference} on {Machine} {Learning}}}. \bibinfo{publisher}{PMLR}, \bibinfo{pages}{2709--2720}.
\newblock
\newblock
\shownote{ISSN: 2640-3498}.


\bibitem[Chase(2022)]%
        {chaseLangChain2022}
\bibfield{author}{\bibinfo{person}{Harrison Chase}.} \bibinfo{year}{2022}\natexlab{}.
\newblock \bibinfo{title}{{LangChain}}.
\newblock
\newblock


\bibitem[Chen et~al\mbox{.}(2019)]%
        {chenHierarchical2019}
\bibfield{author}{\bibinfo{person}{Lele Chen}, \bibinfo{person}{Ross~K. Maddox}, \bibinfo{person}{Zhiyao Duan}, {and} \bibinfo{person}{Chenliang Xu}.} \bibinfo{year}{2019}\natexlab{}.
\newblock \showarticletitle{Hierarchical {Cross}-{Modal} {Talking} {Face} {Generation} {With} {Dynamic} {Pixel}-{Wise} {Loss}}. In \bibinfo{booktitle}{\emph{Proceedings of the {IEEE}/{CVF} {Conference} on {Computer} {Vision} and {Pattern} {Recognition}}}. \bibinfo{pages}{7832--7841}.
\newblock


\bibitem[Choi et~al\mbox{.}(2020)]%
        {choiAttentron2020}
\bibfield{author}{\bibinfo{person}{Seungwoo Choi}, \bibinfo{person}{Seungju Han}, \bibinfo{person}{Dongyoung Kim}, {and} \bibinfo{person}{Sungjoo Ha}.} \bibinfo{year}{2020}\natexlab{}.
\newblock \showarticletitle{Attentron: {Few}-{Shot} {Text}-to-{Speech} {Utilizing} {Attention}-{Based} {Variable}-{Length} {Embedding}}. In \bibinfo{booktitle}{\emph{Interspeech 2020}}. \bibinfo{publisher}{ISCA}, \bibinfo{pages}{2007--2011}.
\newblock


\bibitem[Chollet(2017)]%
        {cholletXception2017}
\bibfield{author}{\bibinfo{person}{Francois Chollet}.} \bibinfo{year}{2017}\natexlab{}.
\newblock \showarticletitle{Xception: {Deep} {Learning} {With} {Depthwise} {Separable} {Convolutions}}. In \bibinfo{booktitle}{\emph{Proceedings of the {IEEE} {Conference} on {Computer} {Vision} and {Pattern} {Recognition}}}. \bibinfo{pages}{1251--1258}.
\newblock


\bibitem[Chugh et~al\mbox{.}(2020)]%
        {chughNot2020}
\bibfield{author}{\bibinfo{person}{Komal Chugh}, \bibinfo{person}{Parul Gupta}, \bibinfo{person}{Abhinav Dhall}, {and} \bibinfo{person}{Ramanathan Subramanian}.} \bibinfo{year}{2020}\natexlab{}.
\newblock \showarticletitle{Not made for each other- {Audio}-{Visual} {Dissonance}-based {Deepfake} {Detection} and {Localization}}. In \bibinfo{booktitle}{\emph{Proceedings of the 28th {ACM} {International} {Conference} on {Multimedia}}} \emph{(\bibinfo{series}{{MM} '20})}. \bibinfo{publisher}{Association for Computing Machinery}, \bibinfo{address}{New York, NY, USA}, \bibinfo{pages}{439--447}.
\newblock
\showISBNx{978-1-4503-7988-5}


\bibitem[Chung et~al\mbox{.}(2018)]%
        {chungVoxCeleb22018}
\bibfield{author}{\bibinfo{person}{Joon~Son Chung}, \bibinfo{person}{Arsha Nagrani}, {and} \bibinfo{person}{Andrew Zisserman}.} \bibinfo{year}{2018}\natexlab{}.
\newblock \showarticletitle{{VoxCeleb2}: {Deep} {Speaker} {Recognition}}. In \bibinfo{booktitle}{\emph{Interspeech 2018}}. \bibinfo{publisher}{ISCA}, \bibinfo{pages}{1086--1090}.
\newblock


\bibitem[Coccomini et~al\mbox{.}(2022)]%
        {coccominiCombining2022}
\bibfield{author}{\bibinfo{person}{Davide~Alessandro Coccomini}, \bibinfo{person}{Nicola Messina}, \bibinfo{person}{Claudio Gennaro}, {and} \bibinfo{person}{Fabrizio Falchi}.} \bibinfo{year}{2022}\natexlab{}.
\newblock \showarticletitle{Combining {EfficientNet} and {Vision} {Transformers} for {Video} {Deepfake} {Detection}}. In \bibinfo{booktitle}{\emph{Image {Analysis} and {Processing} – {ICIAP} 2022}} \emph{(\bibinfo{series}{Lecture {Notes} in {Computer} {Science}})}, \bibfield{editor}{\bibinfo{person}{Stan Sclaroff}, \bibinfo{person}{Cosimo Distante}, \bibinfo{person}{Marco Leo}, \bibinfo{person}{Giovanni~M. Farinella}, {and} \bibinfo{person}{Federico Tombari}} (Eds.). \bibinfo{publisher}{Springer International Publishing}, \bibinfo{address}{Cham}, \bibinfo{pages}{219--229}.
\newblock
\showISBNx{978-3-031-06433-3}


\bibitem[Dolhansky et~al\mbox{.}(2020)]%
        {dolhanskyDeepFake2020}
\bibfield{author}{\bibinfo{person}{Brian Dolhansky}, \bibinfo{person}{Joanna Bitton}, \bibinfo{person}{Ben Pflaum}, \bibinfo{person}{Jikuo Lu}, \bibinfo{person}{Russ Howes}, \bibinfo{person}{Menglin Wang}, {and} \bibinfo{person}{Cristian~Canton Ferrer}.} \bibinfo{year}{2020}\natexlab{}.
\newblock \bibinfo{title}{The {DeepFake} {Detection} {Challenge} ({DFDC}) {Dataset}}.
\newblock
\newblock
\newblock
\shownote{arXiv: 2006.07397 [cs]}.


\bibitem[Défossez et~al\mbox{.}(2020)]%
        {defossezReal2020}
\bibfield{author}{\bibinfo{person}{Alexandre Défossez}, \bibinfo{person}{Gabriel Synnaeve}, {and} \bibinfo{person}{Yossi Adi}.} \bibinfo{year}{2020}\natexlab{}.
\newblock \showarticletitle{Real {Time} {Speech} {Enhancement} in the {Waveform} {Domain}}. In \bibinfo{booktitle}{\emph{Interspeech 2020}}. \bibinfo{address}{Shanghai, China}, \bibinfo{pages}{3291--3295}.
\newblock


\bibitem[Feng et~al\mbox{.}(2023)]%
        {fengSelfSupervised2023}
\bibfield{author}{\bibinfo{person}{Chao Feng}, \bibinfo{person}{Ziyang Chen}, {and} \bibinfo{person}{Andrew Owens}.} \bibinfo{year}{2023}\natexlab{}.
\newblock \showarticletitle{Self-{Supervised} {Video} {Forensics} by {Audio}-{Visual} {Anomaly} {Detection}}. In \bibinfo{booktitle}{\emph{Proceedings of the {IEEE}/{CVF} {Conference} on {Computer} {Vision} and {Pattern} {Recognition}}}. \bibinfo{pages}{10491--10503}.
\newblock


\bibitem[Ge et~al\mbox{.}(2022)]%
        {geLong2022}
\bibfield{author}{\bibinfo{person}{Songwei Ge}, \bibinfo{person}{Thomas Hayes}, \bibinfo{person}{Harry Yang}, \bibinfo{person}{Xi Yin}, \bibinfo{person}{Guan Pang}, \bibinfo{person}{David Jacobs}, \bibinfo{person}{Jia-Bin Huang}, {and} \bibinfo{person}{Devi Parikh}.} \bibinfo{year}{2022}\natexlab{}.
\newblock \showarticletitle{Long {Video} {Generation} with {Time}-{Agnostic} {VQGAN} and {Time}-{Sensitive} {Transformer}}. In \bibinfo{booktitle}{\emph{Proceedings of the {European} {Conference} on {Computer} {Vision} ({ECCV})}} \emph{(\bibinfo{series}{Lecture {Notes} in {Computer} {Science}})}, \bibfield{editor}{\bibinfo{person}{Shai Avidan}, \bibinfo{person}{Gabriel Brostow}, \bibinfo{person}{Moustapha Cissé}, \bibinfo{person}{Giovanni~Maria Farinella}, {and} \bibinfo{person}{Tal Hassner}} (Eds.). \bibinfo{publisher}{Springer Nature Switzerland}, \bibinfo{address}{Cham}, \bibinfo{pages}{102--118}.
\newblock
\showISBNx{978-3-031-19790-1}


\bibitem[Guo et~al\mbox{.}(2021)]%
        {guoADNeRF2021}
\bibfield{author}{\bibinfo{person}{Yudong Guo}, \bibinfo{person}{Keyu Chen}, \bibinfo{person}{Sen Liang}, \bibinfo{person}{Yong-Jin Liu}, \bibinfo{person}{Hujun Bao}, {and} \bibinfo{person}{Juyong Zhang}.} \bibinfo{year}{2021}\natexlab{}.
\newblock \showarticletitle{{AD}-{NeRF}: {Audio} {Driven} {Neural} {Radiance} {Fields} for {Talking} {Head} {Synthesis}}. In \bibinfo{booktitle}{\emph{Proceedings of the {IEEE}/{CVF} {International} {Conference} on {Computer} {Vision}}}. \bibinfo{pages}{5784--5794}.
\newblock


\bibitem[Haliassos et~al\mbox{.}(2021)]%
        {haliassosLips2021}
\bibfield{author}{\bibinfo{person}{Alexandros Haliassos}, \bibinfo{person}{Konstantinos Vougioukas}, \bibinfo{person}{Stavros Petridis}, {and} \bibinfo{person}{Maja Pantic}.} \bibinfo{year}{2021}\natexlab{}.
\newblock \showarticletitle{Lips {Don}'t {Lie}: {A} {Generalisable} and {Robust} {Approach} {To} {Face} {Forgery} {Detection}}. In \bibinfo{booktitle}{\emph{Proceedings of the {IEEE}/{CVF} {Conference} on {Computer} {Vision} and {Pattern} {Recognition}}}. \bibinfo{pages}{5039--5049}.
\newblock


\bibitem[He et~al\mbox{.}(2021)]%
        {heForgeryNet2021}
\bibfield{author}{\bibinfo{person}{Yinan He}, \bibinfo{person}{Bei Gan}, \bibinfo{person}{Siyu Chen}, \bibinfo{person}{Yichun Zhou}, \bibinfo{person}{Guojun Yin}, \bibinfo{person}{Luchuan Song}, \bibinfo{person}{Lu Sheng}, \bibinfo{person}{Jing Shao}, {and} \bibinfo{person}{Ziwei Liu}.} \bibinfo{year}{2021}\natexlab{}.
\newblock \showarticletitle{{ForgeryNet}: {A} {Versatile} {Benchmark} for {Comprehensive} {Forgery} {Analysis}}. In \bibinfo{booktitle}{\emph{Proceedings of the {IEEE}/{CVF} {Conference} on {Computer} {Vision} and {Pattern} {Recognition}}}. \bibinfo{pages}{4360--4369}.
\newblock


\bibitem[Heo et~al\mbox{.}(2023)]%
        {heoDeepFake2023}
\bibfield{author}{\bibinfo{person}{Young-Jin Heo}, \bibinfo{person}{Woon-Ha Yeo}, {and} \bibinfo{person}{Byung-Gyu Kim}.} \bibinfo{year}{2023}\natexlab{}.
\newblock \showarticletitle{{DeepFake} detection algorithm based on improved vision transformer}.
\newblock \bibinfo{journal}{\emph{Applied Intelligence}} \bibinfo{volume}{53}, \bibinfo{number}{7} (\bibinfo{date}{April} \bibinfo{year}{2023}), \bibinfo{pages}{7512--7527}.
\newblock
\showISSN{1573-7497}


\bibitem[Heusel et~al\mbox{.}(2017)]%
        {heuselGANs2017}
\bibfield{author}{\bibinfo{person}{Martin Heusel}, \bibinfo{person}{Hubert Ramsauer}, \bibinfo{person}{Thomas Unterthiner}, \bibinfo{person}{Bernhard Nessler}, {and} \bibinfo{person}{Sepp Hochreiter}.} \bibinfo{year}{2017}\natexlab{}.
\newblock \showarticletitle{{GANs} {Trained} by a {Two} {Time}-{Scale} {Update} {Rule} {Converge} to a {Local} {Nash} {Equilibrium}}. In \bibinfo{booktitle}{\emph{Advances in {Neural} {Information} {Processing} {Systems}}}, Vol.~\bibinfo{volume}{30}. \bibinfo{publisher}{Curran Associates, Inc.}
\newblock


\bibitem[Ilyas et~al\mbox{.}(2023)]%
        {ilyasAVFakeNet2023}
\bibfield{author}{\bibinfo{person}{Hafsa Ilyas}, \bibinfo{person}{Ali Javed}, {and} \bibinfo{person}{Khalid~Mahmood Malik}.} \bibinfo{year}{2023}\natexlab{}.
\newblock \showarticletitle{{AVFakeNet}: {A} unified end-to-end {Dense} {Swin} {Transformer} deep learning model for audio–visual deepfakes detection}.
\newblock \bibinfo{journal}{\emph{Applied Soft Computing}}  \bibinfo{volume}{136} (\bibinfo{date}{March} \bibinfo{year}{2023}), \bibinfo{pages}{110124}.
\newblock
\showISSN{1568-4946}


\bibitem[Ji et~al\mbox{.}(2022)]%
        {jiEAMM2022}
\bibfield{author}{\bibinfo{person}{Xinya Ji}, \bibinfo{person}{Hang Zhou}, \bibinfo{person}{Kaisiyuan Wang}, \bibinfo{person}{Qianyi Wu}, \bibinfo{person}{Wayne Wu}, \bibinfo{person}{Feng Xu}, {and} \bibinfo{person}{Xun Cao}.} \bibinfo{year}{2022}\natexlab{}.
\newblock \showarticletitle{{EAMM}: {One}-{Shot} {Emotional} {Talking} {Face} via {Audio}-{Based} {Emotion}-{Aware} {Motion} {Model}}. In \bibinfo{booktitle}{\emph{{ACM} {SIGGRAPH} 2022 {Conference} {Proceedings}}} \emph{(\bibinfo{series}{{SIGGRAPH} '22})}. \bibinfo{publisher}{Association for Computing Machinery}, \bibinfo{address}{New York, NY, USA}, \bibinfo{pages}{1--10}.
\newblock
\showISBNx{978-1-4503-9337-9}


\bibitem[Jia et~al\mbox{.}(2018)]%
        {jiaTransfer2018}
\bibfield{author}{\bibinfo{person}{Ye Jia}, \bibinfo{person}{Yu Zhang}, \bibinfo{person}{Ron~J. Weiss}, \bibinfo{person}{Quan Wang}, \bibinfo{person}{Jonathan Shen}, \bibinfo{person}{Fei Ren}, \bibinfo{person}{Zhifeng Chen}, \bibinfo{person}{Patrick Nguyen}, \bibinfo{person}{Ruoming Pang}, \bibinfo{person}{Ignacio~Lopez Moreno}, {and} \bibinfo{person}{Yonghui Wu}.} \bibinfo{year}{2018}\natexlab{}.
\newblock \showarticletitle{Transfer learning from speaker verification to multispeaker text-to-speech synthesis}. In \bibinfo{booktitle}{\emph{Proceedings of the 32nd {International} {Conference} on {Neural} {Information} {Processing} {Systems}}} \emph{(\bibinfo{series}{{NIPS}'18})}. \bibinfo{publisher}{Curran Associates Inc.}, \bibinfo{address}{Red Hook, NY, USA}, \bibinfo{pages}{4485--4495}.
\newblock


\bibitem[Jiang et~al\mbox{.}(2020)]%
        {jiangDeeperForensics12020}
\bibfield{author}{\bibinfo{person}{Liming Jiang}, \bibinfo{person}{Ren Li}, \bibinfo{person}{Wayne Wu}, \bibinfo{person}{Chen Qian}, {and} \bibinfo{person}{Chen~Change Loy}.} \bibinfo{year}{2020}\natexlab{}.
\newblock \showarticletitle{{DeeperForensics}-1.0: {A} {Large}-{Scale} {Dataset} for {Real}-{World} {Face} {Forgery} {Detection}}. In \bibinfo{booktitle}{\emph{Proceedings of the {IEEE}/{CVF} {Conference} on {Computer} {Vision} and {Pattern} {Recognition}}}. \bibinfo{pages}{2889--2898}.
\newblock


\bibitem[Jiang et~al\mbox{.}(2023a)]%
        {jiangMegaTTS2023a}
\bibfield{author}{\bibinfo{person}{Ziyue Jiang}, \bibinfo{person}{Jinglin Liu}, \bibinfo{person}{Yi Ren}, \bibinfo{person}{Jinzheng He}, \bibinfo{person}{Chen Zhang}, \bibinfo{person}{Zhenhui Ye}, \bibinfo{person}{Pengfei Wei}, \bibinfo{person}{Chunfeng Wang}, \bibinfo{person}{Xiang Yin}, \bibinfo{person}{Zejun Ma}, {and} \bibinfo{person}{Zhou Zhao}.} \bibinfo{year}{2023}\natexlab{a}.
\newblock \bibinfo{title}{Mega-{TTS} 2: {Zero}-{Shot} {Text}-to-{Speech} with {Arbitrary} {Length} {Speech} {Prompts}}.
\newblock
\newblock
\newblock
\shownote{arXiv:2307.07218 [cs, eess]}.


\bibitem[Jiang et~al\mbox{.}(2023b)]%
        {jiangMegaTTS2023}
\bibfield{author}{\bibinfo{person}{Ziyue Jiang}, \bibinfo{person}{Yi Ren}, \bibinfo{person}{Zhenhui Ye}, \bibinfo{person}{Jinglin Liu}, \bibinfo{person}{Chen Zhang}, \bibinfo{person}{Qian Yang}, \bibinfo{person}{Shengpeng Ji}, \bibinfo{person}{Rongjie Huang}, \bibinfo{person}{Chunfeng Wang}, \bibinfo{person}{Xiang Yin}, \bibinfo{person}{Zejun Ma}, {and} \bibinfo{person}{Zhou Zhao}.} \bibinfo{year}{2023}\natexlab{b}.
\newblock \bibinfo{title}{Mega-{TTS}: {Zero}-{Shot} {Text}-to-{Speech} at {Scale} with {Intrinsic} {Inductive} {Bias}}.
\newblock
\newblock
\newblock
\shownote{arXiv:2306.03509 [cs, eess]}.


\bibitem[Khalid et~al\mbox{.}(2021)]%
        {khalidFakeAVCeleb2021}
\bibfield{author}{\bibinfo{person}{Hasam Khalid}, \bibinfo{person}{Shahroz Tariq}, {and} \bibinfo{person}{Simon~S. Woo}.} \bibinfo{year}{2021}\natexlab{}.
\newblock \bibinfo{title}{{FakeAVCeleb}: {A} {Novel} {Audio}-{Video} {Multimodal} {Deepfake} {Dataset}}.
\newblock
\newblock
\newblock
\shownote{arXiv: 2108.05080 [cs]}.


\bibitem[Kilgour et~al\mbox{.}(2019)]%
        {kilgourFrechet2019}
\bibfield{author}{\bibinfo{person}{Kevin Kilgour}, \bibinfo{person}{Mauricio Zuluaga}, \bibinfo{person}{Dominik Roblek}, {and} \bibinfo{person}{Matthew Sharifi}.} \bibinfo{year}{2019}\natexlab{}.
\newblock \bibinfo{title}{Fréchet {Audio} {Distance}: {A} {Metric} for {Evaluating} {Music} {Enhancement} {Algorithms}}.
\newblock
\newblock
\newblock
\shownote{arXiv:1812.08466 [cs, eess]}.


\bibitem[Kim et~al\mbox{.}(2021)]%
        {kimConditional2021}
\bibfield{author}{\bibinfo{person}{Jaehyeon Kim}, \bibinfo{person}{Jungil Kong}, {and} \bibinfo{person}{Juhee Son}.} \bibinfo{year}{2021}\natexlab{}.
\newblock \showarticletitle{Conditional {Variational} {Autoencoder} with {Adversarial} {Learning} for {End}-to-{End} {Text}-to-{Speech}}. In \bibinfo{booktitle}{\emph{Proceedings of the 38th {International} {Conference} on {Machine} {Learning}}}. \bibinfo{publisher}{PMLR}, \bibinfo{pages}{5530--5540}.
\newblock
\newblock
\shownote{ISSN: 2640-3498}.


\bibitem[Korshunov and Marcel(2018)]%
        {korshunovDeepFakes2018}
\bibfield{author}{\bibinfo{person}{Pavel Korshunov} {and} \bibinfo{person}{Sebastien Marcel}.} \bibinfo{year}{2018}\natexlab{}.
\newblock \bibinfo{title}{{DeepFakes}: a {New} {Threat} to {Face} {Recognition}? {Assessment} and {Detection}}.
\newblock
\newblock
\newblock
\shownote{arXiv:1812.08685 [cs]}.


\bibitem[Kwon et~al\mbox{.}(2021)]%
        {kwonKoDF2021}
\bibfield{author}{\bibinfo{person}{Patrick Kwon}, \bibinfo{person}{Jaeseong You}, \bibinfo{person}{Gyuhyeon Nam}, \bibinfo{person}{Sungwoo Park}, {and} \bibinfo{person}{Gyeongsu Chae}.} \bibinfo{year}{2021}\natexlab{}.
\newblock \showarticletitle{{KoDF}: {A} {Large}-{Scale} {Korean} {DeepFake} {Detection} {Dataset}}. In \bibinfo{booktitle}{\emph{Proceedings of the {IEEE}/{CVF} {International} {Conference} on {Computer} {Vision}}}. \bibinfo{pages}{10744--10753}.
\newblock


\bibitem[Li et~al\mbox{.}(2020a)]%
        {liFace2020}
\bibfield{author}{\bibinfo{person}{Lingzhi Li}, \bibinfo{person}{Jianmin Bao}, \bibinfo{person}{Ting Zhang}, \bibinfo{person}{Hao Yang}, \bibinfo{person}{Dong Chen}, \bibinfo{person}{Fang Wen}, {and} \bibinfo{person}{Baining Guo}.} \bibinfo{year}{2020}\natexlab{a}.
\newblock \showarticletitle{Face {X}-{Ray} for {More} {General} {Face} {Forgery} {Detection}}. In \bibinfo{booktitle}{\emph{Proceedings of the {IEEE}/{CVF} {Conference} on {Computer} {Vision} and {Pattern} {Recognition}}}. \bibinfo{pages}{5001--5010}.
\newblock


\bibitem[Li and Lyu(2019)]%
        {liExposing2019}
\bibfield{author}{\bibinfo{person}{Yuezun Li} {and} \bibinfo{person}{Siwei Lyu}.} \bibinfo{year}{2019}\natexlab{}.
\newblock \showarticletitle{Exposing {DeepFake} {Videos} {By} {Detecting} {Face} {Warping} {Artifacts}}. In \bibinfo{booktitle}{\emph{Proceedings of the {IEEE}/{CVF} {Conference} on {Computer} {Vision} and {Pattern} {Recognition} {Workshops}}}. \bibinfo{pages}{7}.
\newblock


\bibitem[Li et~al\mbox{.}(2020b)]%
        {liCelebDF2020}
\bibfield{author}{\bibinfo{person}{Yuezun Li}, \bibinfo{person}{Xin Yang}, \bibinfo{person}{Pu Sun}, \bibinfo{person}{Honggang Qi}, {and} \bibinfo{person}{Siwei Lyu}.} \bibinfo{year}{2020}\natexlab{b}.
\newblock \showarticletitle{Celeb-{DF}: {A} {Large}-{Scale} {Challenging} {Dataset} for {DeepFake} {Forensics}}. In \bibinfo{booktitle}{\emph{Proceedings of the {IEEE}/{CVF} {Conference} on {Computer} {Vision} and {Pattern} {Recognition}}}. \bibinfo{pages}{3207--3216}.
\newblock


\bibitem[Liu et~al\mbox{.}(2023)]%
        {liuASVspoof2023}
\bibfield{author}{\bibinfo{person}{Xuechen Liu}, \bibinfo{person}{Xin Wang}, \bibinfo{person}{Md Sahidullah}, \bibinfo{person}{Jose Patino}, \bibinfo{person}{Héctor Delgado}, \bibinfo{person}{Tomi Kinnunen}, \bibinfo{person}{Massimiliano Todisco}, \bibinfo{person}{Junichi Yamagishi}, \bibinfo{person}{Nicholas Evans}, \bibinfo{person}{Andreas Nautsch}, {and} \bibinfo{person}{Kong~Aik Lee}.} \bibinfo{year}{2023}\natexlab{}.
\newblock \showarticletitle{{ASVspoof} 2021: {Towards} {Spoofed} and {Deepfake} {Speech} {Detection} in the {Wild}}.
\newblock \bibinfo{journal}{\emph{IEEE/ACM Transactions on Audio, Speech, and Language Processing}}  \bibinfo{volume}{31} (\bibinfo{year}{2023}), \bibinfo{pages}{2507--2522}.
\newblock
\showISSN{2329-9304}


\bibitem[Mittal et~al\mbox{.}(2020)]%
        {mittalEmotions2020}
\bibfield{author}{\bibinfo{person}{Trisha Mittal}, \bibinfo{person}{Uttaran Bhattacharya}, \bibinfo{person}{Rohan Chandra}, \bibinfo{person}{Aniket Bera}, {and} \bibinfo{person}{Dinesh Manocha}.} \bibinfo{year}{2020}\natexlab{}.
\newblock \showarticletitle{Emotions {Don}'t {Lie}: {An} {Audio}-{Visual} {Deepfake} {Detection} {Method} using {Affective} {Cues}}. In \bibinfo{booktitle}{\emph{Proceedings of the 28th {ACM} {International} {Conference} on {Multimedia}}} \emph{(\bibinfo{series}{{MM} '20})}. \bibinfo{publisher}{Association for Computing Machinery}, \bibinfo{address}{New York, NY, USA}, \bibinfo{pages}{2823--2832}.
\newblock
\showISBNx{978-1-4503-7988-5}


\bibitem[Narayan et~al\mbox{.}(2023)]%
        {narayanDFPlatter2023}
\bibfield{author}{\bibinfo{person}{Kartik Narayan}, \bibinfo{person}{Harsh Agarwal}, \bibinfo{person}{Kartik Thakral}, \bibinfo{person}{Surbhi Mittal}, \bibinfo{person}{Mayank Vatsa}, {and} \bibinfo{person}{Richa Singh}.} \bibinfo{year}{2023}\natexlab{}.
\newblock \showarticletitle{{DF}-{Platter}: {Multi}-{Face} {Heterogeneous} {Deepfake} {Dataset}}. In \bibinfo{booktitle}{\emph{Proceedings of the {IEEE}/{CVF} {Conference} on {Computer} {Vision} and {Pattern} {Recognition}}}. \bibinfo{pages}{9739--9748}.
\newblock


\bibitem[Nick and Andrew(2019)]%
        {nickContributing2019}
\bibfield{author}{\bibinfo{person}{Dufou Nick} {and} \bibinfo{person}{Jigsaw Andrew}.} \bibinfo{year}{2019}\natexlab{}.
\newblock \bibinfo{title}{Contributing {Data} to {Deepfake} {Detection} {Research}}.
\newblock
\newblock


\bibitem[Niizumi et~al\mbox{.}(2021)]%
        {niizumiBYOL2021}
\bibfield{author}{\bibinfo{person}{Daisuke Niizumi}, \bibinfo{person}{Daiki Takeuchi}, \bibinfo{person}{Yasunori Ohishi}, \bibinfo{person}{Noboru Harada}, {and} \bibinfo{person}{Kunio Kashino}.} \bibinfo{year}{2021}\natexlab{}.
\newblock \showarticletitle{{BYOL} for {Audio}: {Self}-{Supervised} {Learning} for {General}-{Purpose} {Audio} {Representation}}. In \bibinfo{booktitle}{\emph{2021 {International} {Joint} {Conference} on {Neural} {Networks} ({IJCNN})}}. \bibinfo{pages}{1--8}.
\newblock
\newblock
\shownote{ISSN: 2161-4407}.


\bibitem[Oorloff et~al\mbox{.}(2024)]%
        {oorloffAVFF2024}
\bibfield{author}{\bibinfo{person}{Trevine Oorloff}, \bibinfo{person}{Surya Koppisetti}, \bibinfo{person}{Nicolò Bonettini}, \bibinfo{person}{Divyaraj Solanki}, \bibinfo{person}{Ben Colman}, \bibinfo{person}{Yaser Yacoob}, \bibinfo{person}{Ali Shahriyari}, {and} \bibinfo{person}{Gaurav Bharaj}.} \bibinfo{year}{2024}\natexlab{}.
\newblock \showarticletitle{{AVFF}: {Audio}-{Visual} {Feature} {Fusion} for {Video} {Deepfake} {Detection}}. In \bibinfo{booktitle}{\emph{Proceedings of the {IEEE}/{CVF} {Conference} on {Computer} {Vision} and {Pattern} {Recognition}}}. \bibinfo{pages}{27102--27112}.
\newblock


\bibitem[Plaquet and Bredin(2023)]%
        {plaquetPowerset2023}
\bibfield{author}{\bibinfo{person}{Alexis Plaquet} {and} \bibinfo{person}{Hervé Bredin}.} \bibinfo{year}{2023}\natexlab{}.
\newblock \showarticletitle{Powerset multi-class cross entropy loss for neural speaker diarization}. In \bibinfo{booktitle}{\emph{{INTERSPEECH} 2023}}. \bibinfo{publisher}{ISCA}, \bibinfo{pages}{3222--3226}.
\newblock


\bibitem[Prajwal et~al\mbox{.}(2020)]%
        {prajwalLip2020}
\bibfield{author}{\bibinfo{person}{K~R Prajwal}, \bibinfo{person}{Rudrabha Mukhopadhyay}, \bibinfo{person}{Vinay~P. Namboodiri}, {and} \bibinfo{person}{C.V. Jawahar}.} \bibinfo{year}{2020}\natexlab{}.
\newblock \showarticletitle{A {Lip} {Sync} {Expert} {Is} {All} {You} {Need} for {Speech} to {Lip} {Generation} {In} the {Wild}}. In \bibinfo{booktitle}{\emph{Proceedings of the 28th {ACM} {International} {Conference} on {Multimedia}}} \emph{(\bibinfo{series}{{MM} '20})}. \bibinfo{publisher}{Association for Computing Machinery}, \bibinfo{address}{New York, NY, USA}, \bibinfo{pages}{484--492}.
\newblock
\showISBNx{978-1-4503-7988-5}


\bibitem[Qian et~al\mbox{.}(2020)]%
        {qianThinking2020}
\bibfield{author}{\bibinfo{person}{Yuyang Qian}, \bibinfo{person}{Guojun Yin}, \bibinfo{person}{Lu Sheng}, \bibinfo{person}{Zixuan Chen}, {and} \bibinfo{person}{Jing Shao}.} \bibinfo{year}{2020}\natexlab{}.
\newblock \showarticletitle{Thinking in {Frequency}: {Face} {Forgery} {Detection} by {Mining} {Frequency}-{Aware} {Clues}}. In \bibinfo{booktitle}{\emph{Proceedings of the {European} {Conference} on {Computer} {Vision} ({ECCV})}} \emph{(\bibinfo{series}{Lecture {Notes} in {Computer} {Science}})}, \bibfield{editor}{\bibinfo{person}{Andrea Vedaldi}, \bibinfo{person}{Horst Bischof}, \bibinfo{person}{Thomas Brox}, {and} \bibinfo{person}{Jan-Michael Frahm}} (Eds.). \bibinfo{publisher}{Springer International Publishing}, \bibinfo{address}{Cham}, \bibinfo{pages}{86--103}.
\newblock
\showISBNx{978-3-030-58610-2}


\bibitem[Radford et~al\mbox{.}(2023)]%
        {radfordRobust2023}
\bibfield{author}{\bibinfo{person}{Alec Radford}, \bibinfo{person}{Jong~Wook Kim}, \bibinfo{person}{Tao Xu}, \bibinfo{person}{Greg Brockman}, \bibinfo{person}{Christine Mcleavey}, {and} \bibinfo{person}{Ilya Sutskever}.} \bibinfo{year}{2023}\natexlab{}.
\newblock \showarticletitle{Robust {Speech} {Recognition} via {Large}-{Scale} {Weak} {Supervision}}. In \bibinfo{booktitle}{\emph{Proceedings of the 40th {International} {Conference} on {Machine} {Learning}}}. \bibinfo{publisher}{PMLR}, \bibinfo{pages}{28492--28518}.
\newblock
\newblock
\shownote{ISSN: 2640-3498}.


\bibitem[Raza and Malik(2023)]%
        {razaMultimodaltrace2023}
\bibfield{author}{\bibinfo{person}{Muhammad~Anas Raza} {and} \bibinfo{person}{Khalid~Mahmood Malik}.} \bibinfo{year}{2023}\natexlab{}.
\newblock \showarticletitle{Multimodaltrace: {Deepfake} {Detection} {Using} {Audiovisual} {Representation} {Learning}}. In \bibinfo{booktitle}{\emph{Proceedings of the {IEEE}/{CVF} {Conference} on {Computer} {Vision} and {Pattern} {Recognition}}}. \bibinfo{pages}{993--1000}.
\newblock


\bibitem[Rossler et~al\mbox{.}(2019)]%
        {rosslerFaceForensics2019}
\bibfield{author}{\bibinfo{person}{Andreas Rossler}, \bibinfo{person}{Davide Cozzolino}, \bibinfo{person}{Luisa Verdoliva}, \bibinfo{person}{Christian Riess}, \bibinfo{person}{Justus Thies}, {and} \bibinfo{person}{Matthias Niessner}.} \bibinfo{year}{2019}\natexlab{}.
\newblock \showarticletitle{{FaceForensics}++: {Learning} to {Detect} {Manipulated} {Facial} {Images}}. In \bibinfo{booktitle}{\emph{Proceedings of the {IEEE}/{CVF} {International} {Conference} on {Computer} {Vision}}}. \bibinfo{pages}{1--11}.
\newblock


\bibitem[Shahzad et~al\mbox{.}(2022)]%
        {shahzadLip2022}
\bibfield{author}{\bibinfo{person}{Sahibzada~Adil Shahzad}, \bibinfo{person}{Ammarah Hashmi}, \bibinfo{person}{Sarwar Khan}, \bibinfo{person}{Yan-Tsung Peng}, \bibinfo{person}{Yu Tsao}, {and} \bibinfo{person}{Hsin-Min Wang}.} \bibinfo{year}{2022}\natexlab{}.
\newblock \showarticletitle{Lip {Sync} {Matters}: {A} {Novel} {Multimodal} {Forgery} {Detector}}. In \bibinfo{booktitle}{\emph{2022 {Asia}-{Pacific} {Signal} and {Information} {Processing} {Association} {Annual} {Summit} and {Conference} ({APSIPA} {ASC})}}. \bibinfo{pages}{1885--1892}.
\newblock
\newblock
\shownote{ISSN: 2640-0103}.


\bibitem[Shao et~al\mbox{.}(2024)]%
        {shaoDetecting2024}
\bibfield{author}{\bibinfo{person}{Rui Shao}, \bibinfo{person}{Tianxing Wu}, \bibinfo{person}{Jianlong Wu}, \bibinfo{person}{Liqiang Nie}, {and} \bibinfo{person}{Ziwei Liu}.} \bibinfo{year}{2024}\natexlab{}.
\newblock \showarticletitle{Detecting and {Grounding} {Multi}-{Modal} {Media} {Manipulation} and {Beyond}}.
\newblock \bibinfo{journal}{\emph{IEEE Transactions on Pattern Analysis and Machine Intelligence}} (\bibinfo{year}{2024}), \bibinfo{pages}{1--18}.
\newblock
\showISSN{1939-3539}


\bibitem[Shen et~al\mbox{.}(2023a)]%
        {shenNaturalSpeech2023}
\bibfield{author}{\bibinfo{person}{Kai Shen}, \bibinfo{person}{Zeqian Ju}, \bibinfo{person}{Xu Tan}, \bibinfo{person}{Yanqing Liu}, \bibinfo{person}{Yichong Leng}, \bibinfo{person}{Lei He}, \bibinfo{person}{Tao Qin}, \bibinfo{person}{Sheng Zhao}, {and} \bibinfo{person}{Jiang Bian}.} \bibinfo{year}{2023}\natexlab{a}.
\newblock \bibinfo{title}{{NaturalSpeech} 2: {Latent} {Diffusion} {Models} are {Natural} and {Zero}-{Shot} {Speech} and {Singing} {Synthesizers}}.
\newblock
\newblock
\newblock
\shownote{arXiv:2304.09116 [cs, eess]}.


\bibitem[Shen et~al\mbox{.}(2023b)]%
        {shenDiffTalk2023}
\bibfield{author}{\bibinfo{person}{Shuai Shen}, \bibinfo{person}{Wenliang Zhao}, \bibinfo{person}{Zibin Meng}, \bibinfo{person}{Wanhua Li}, \bibinfo{person}{Zheng Zhu}, \bibinfo{person}{Jie Zhou}, {and} \bibinfo{person}{Jiwen Lu}.} \bibinfo{year}{2023}\natexlab{b}.
\newblock \showarticletitle{{DiffTalk}: {Crafting} {Diffusion} {Models} for {Generalized} {Audio}-{Driven} {Portraits} {Animation}}. In \bibinfo{booktitle}{\emph{Proceedings of the {IEEE}/{CVF} {Conference} on {Computer} {Vision} and {Pattern} {Recognition}}}. \bibinfo{pages}{1982--1991}.
\newblock


\bibitem[Shi et~al\mbox{.}(2023)]%
        {shiTriDet2023}
\bibfield{author}{\bibinfo{person}{Dingfeng Shi}, \bibinfo{person}{Yujie Zhong}, \bibinfo{person}{Qiong Cao}, \bibinfo{person}{Lin Ma}, \bibinfo{person}{Jia Li}, {and} \bibinfo{person}{Dacheng Tao}.} \bibinfo{year}{2023}\natexlab{}.
\newblock \showarticletitle{{TriDet}: {Temporal} {Action} {Detection} {With} {Relative} {Boundary} {Modeling}}. In \bibinfo{booktitle}{\emph{Proceedings of the {IEEE}/{CVF} {Conference} on {Computer} {Vision} and {Pattern} {Recognition}}}. \bibinfo{pages}{18857--18866}.
\newblock


\bibitem[Shiohara and Yamasaki(2022)]%
        {shioharaDetecting2022}
\bibfield{author}{\bibinfo{person}{Kaede Shiohara} {and} \bibinfo{person}{Toshihiko Yamasaki}.} \bibinfo{year}{2022}\natexlab{}.
\newblock \showarticletitle{Detecting {Deepfakes} {With} {Self}-{Blended} {Images}}. In \bibinfo{booktitle}{\emph{Proceedings of the {IEEE}/{CVF} {Conference} on {Computer} {Vision} and {Pattern} {Recognition}}}. \bibinfo{pages}{18720--18729}.
\newblock


\bibitem[Singer et~al\mbox{.}(2022)]%
        {singerMakeAVideo2022}
\bibfield{author}{\bibinfo{person}{Uriel Singer}, \bibinfo{person}{Adam Polyak}, \bibinfo{person}{Thomas Hayes}, \bibinfo{person}{Xi Yin}, \bibinfo{person}{Jie An}, \bibinfo{person}{Songyang Zhang}, \bibinfo{person}{Qiyuan Hu}, \bibinfo{person}{Harry Yang}, \bibinfo{person}{Oron Ashual}, \bibinfo{person}{Oran Gafni}, \bibinfo{person}{Devi Parikh}, \bibinfo{person}{Sonal Gupta}, {and} \bibinfo{person}{Yaniv Taigman}.} \bibinfo{year}{2022}\natexlab{}.
\newblock \bibinfo{title}{Make-{A}-{Video}: {Text}-to-{Video} {Generation} without {Text}-{Video} {Data}}.
\newblock
\newblock
\newblock
\shownote{arXiv:2209.14792 [cs]}.


\bibitem[Tomar(2006)]%
        {tomarConverting2006}
\bibfield{author}{\bibinfo{person}{Suramya Tomar}.} \bibinfo{year}{2006}\natexlab{}.
\newblock \showarticletitle{Converting video formats with {FFmpeg}}.
\newblock \bibinfo{journal}{\emph{Linux Journal}} \bibinfo{volume}{2006}, \bibinfo{number}{146} (\bibinfo{date}{June} \bibinfo{year}{2006}), \bibinfo{pages}{10}.
\newblock
\showISSN{1075-3583}


\bibitem[Touvron et~al\mbox{.}(2023)]%
        {touvronLLaMA2023}
\bibfield{author}{\bibinfo{person}{Hugo Touvron}, \bibinfo{person}{Thibaut Lavril}, \bibinfo{person}{Gautier Izacard}, \bibinfo{person}{Xavier Martinet}, \bibinfo{person}{Marie-Anne Lachaux}, \bibinfo{person}{Timothée Lacroix}, \bibinfo{person}{Baptiste Rozière}, \bibinfo{person}{Naman Goyal}, \bibinfo{person}{Eric Hambro}, \bibinfo{person}{Faisal Azhar}, \bibinfo{person}{Aurelien Rodriguez}, \bibinfo{person}{Armand Joulin}, \bibinfo{person}{Edouard Grave}, {and} \bibinfo{person}{Guillaume Lample}.} \bibinfo{year}{2023}\natexlab{}.
\newblock \bibinfo{title}{{LLaMA}: {Open} and {Efficient} {Foundation} {Language} {Models}}.
\newblock
\newblock
\newblock
\shownote{arXiv:2302.13971 [cs]}.


\bibitem[Wan et~al\mbox{.}(2018)]%
        {wanGeneralized2018}
\bibfield{author}{\bibinfo{person}{Li Wan}, \bibinfo{person}{Quan Wang}, \bibinfo{person}{Alan Papir}, {and} \bibinfo{person}{Ignacio~Lopez Moreno}.} \bibinfo{year}{2018}\natexlab{}.
\newblock \showarticletitle{Generalized {End}-to-{End} {Loss} for {Speaker} {Verification}}. In \bibinfo{booktitle}{\emph{{IEEE} {International} {Conference} on {Acoustics}, {Speech} and {Signal} {Processing} ({ICASSP})}}. \bibinfo{pages}{4879--4883}.
\newblock
\newblock
\shownote{ISSN: 2379-190X}.


\bibitem[Wang et~al\mbox{.}(2024)]%
        {wangExploiting2024}
\bibfield{author}{\bibinfo{person}{Jiazhen Wang}, \bibinfo{person}{Bin Liu}, \bibinfo{person}{Changtao Miao}, \bibinfo{person}{Zhiwei Zhao}, \bibinfo{person}{Wanyi Zhuang}, \bibinfo{person}{Qi Chu}, {and} \bibinfo{person}{Nenghai Yu}.} \bibinfo{year}{2024}\natexlab{}.
\newblock \showarticletitle{Exploiting {Modality}-{Specific} {Features} for {Multi}-{Modal} {Manipulation} {Detection} and {Grounding}}. In \bibinfo{booktitle}{\emph{{IEEE} {International} {Conference} on {Acoustics}, {Speech} and {Signal} {Processing} ({ICASSP})}}. \bibinfo{pages}{4935--4939}.
\newblock
\newblock
\shownote{ISSN: 2379-190X}.


\bibitem[Wang et~al\mbox{.}(2023b)]%
        {wangSeeing2023}
\bibfield{author}{\bibinfo{person}{Jiadong Wang}, \bibinfo{person}{Xinyuan Qian}, \bibinfo{person}{Malu Zhang}, \bibinfo{person}{Robby~T. Tan}, {and} \bibinfo{person}{Haizhou Li}.} \bibinfo{year}{2023}\natexlab{b}.
\newblock \showarticletitle{Seeing {What} {You} {Said}: {Talking} {Face} {Generation} {Guided} by a {Lip} {Reading} {Expert}}. In \bibinfo{booktitle}{\emph{Proceedings of the {IEEE}/{CVF} {Conference} on {Computer} {Vision} and {Pattern} {Recognition}}}. \bibinfo{pages}{14653--14662}.
\newblock


\bibitem[Wang et~al\mbox{.}(2022b)]%
        {wangM2TR2022}
\bibfield{author}{\bibinfo{person}{Junke Wang}, \bibinfo{person}{Zuxuan Wu}, \bibinfo{person}{Wenhao Ouyang}, \bibinfo{person}{Xintong Han}, \bibinfo{person}{Jingjing Chen}, \bibinfo{person}{Yu-Gang Jiang}, {and} \bibinfo{person}{Ser-Nam Li}.} \bibinfo{year}{2022}\natexlab{b}.
\newblock \showarticletitle{{M2TR}: {Multi}-modal {Multi}-scale {Transformers} for {Deepfake} {Detection}}. In \bibinfo{booktitle}{\emph{Proceedings of the 2022 {International} {Conference} on {Multimedia} {Retrieval}}} \emph{(\bibinfo{series}{{ICMR} '22})}. \bibinfo{publisher}{Association for Computing Machinery}, \bibinfo{address}{New York, NY, USA}, \bibinfo{pages}{615--623}.
\newblock
\showISBNx{978-1-4503-9238-9}


\bibitem[Wang et~al\mbox{.}(2023a)]%
        {wangVideoMAE2023}
\bibfield{author}{\bibinfo{person}{Limin Wang}, \bibinfo{person}{Bingkun Huang}, \bibinfo{person}{Zhiyu Zhao}, \bibinfo{person}{Zhan Tong}, \bibinfo{person}{Yinan He}, \bibinfo{person}{Yi Wang}, \bibinfo{person}{Yali Wang}, {and} \bibinfo{person}{Yu Qiao}.} \bibinfo{year}{2023}\natexlab{a}.
\newblock \showarticletitle{{VideoMAE} {V2}: {Scaling} {Video} {Masked} {Autoencoders} {With} {Dual} {Masking}}. In \bibinfo{booktitle}{\emph{Proceedings of the {IEEE}/{CVF} {Conference} on {Computer} {Vision} and {Pattern} {Recognition}}}. \bibinfo{pages}{14549--14560}.
\newblock


\bibitem[Wang et~al\mbox{.}(2022a)]%
        {wangInternVideo2022}
\bibfield{author}{\bibinfo{person}{Yi Wang}, \bibinfo{person}{Kunchang Li}, \bibinfo{person}{Yizhuo Li}, \bibinfo{person}{Yinan He}, \bibinfo{person}{Bingkun Huang}, \bibinfo{person}{Zhiyu Zhao}, \bibinfo{person}{Hongjie Zhang}, \bibinfo{person}{Jilan Xu}, \bibinfo{person}{Yi Liu}, \bibinfo{person}{Zun Wang}, \bibinfo{person}{Sen Xing}, \bibinfo{person}{Guo Chen}, \bibinfo{person}{Junting Pan}, \bibinfo{person}{Jiashuo Yu}, \bibinfo{person}{Yali Wang}, \bibinfo{person}{Limin Wang}, {and} \bibinfo{person}{Yu Qiao}.} \bibinfo{year}{2022}\natexlab{a}.
\newblock \bibinfo{title}{{InternVideo}: {General} {Video} {Foundation} {Models} via {Generative} and {Discriminative} {Learning}}.
\newblock
\newblock
\newblock
\shownote{arXiv:2212.03191 [cs]}.


\bibitem[Wang et~al\mbox{.}(2004)]%
        {wangImage2004}
\bibfield{author}{\bibinfo{person}{Zhou Wang}, \bibinfo{person}{A.C. Bovik}, \bibinfo{person}{H.R. Sheikh}, {and} \bibinfo{person}{E.P. Simoncelli}.} \bibinfo{year}{2004}\natexlab{}.
\newblock \showarticletitle{Image quality assessment: from error visibility to structural similarity}.
\newblock \bibinfo{journal}{\emph{IEEE Transactions on Image Processing}} \bibinfo{volume}{13}, \bibinfo{number}{4} (\bibinfo{date}{April} \bibinfo{year}{2004}), \bibinfo{pages}{600--612}.
\newblock
\showISSN{1941-0042}


\bibitem[Wu et~al\mbox{.}(2023b)]%
        {wuTuneAVideo2023}
\bibfield{author}{\bibinfo{person}{Jay~Zhangjie Wu}, \bibinfo{person}{Yixiao Ge}, \bibinfo{person}{Xintao Wang}, \bibinfo{person}{Stan~Weixian Lei}, \bibinfo{person}{Yuchao Gu}, \bibinfo{person}{Yufei Shi}, \bibinfo{person}{Wynne Hsu}, \bibinfo{person}{Ying Shan}, \bibinfo{person}{Xiaohu Qie}, {and} \bibinfo{person}{Mike~Zheng Shou}.} \bibinfo{year}{2023}\natexlab{b}.
\newblock \showarticletitle{Tune-{A}-{Video}: {One}-{Shot} {Tuning} of {Image} {Diffusion} {Models} for {Text}-to-{Video} {Generation}}. In \bibinfo{booktitle}{\emph{Proceedings of the {IEEE}/{CVF} {International} {Conference} on {Computer} {Vision}}}. \bibinfo{pages}{7623--7633}.
\newblock


\bibitem[Wu et~al\mbox{.}(2023a)]%
        {wuLargeScale2023}
\bibfield{author}{\bibinfo{person}{Yusong Wu}, \bibinfo{person}{Ke Chen}, \bibinfo{person}{Tianyu Zhang}, \bibinfo{person}{Yuchen Hui}, \bibinfo{person}{Taylor Berg-Kirkpatrick}, {and} \bibinfo{person}{Shlomo Dubnov}.} \bibinfo{year}{2023}\natexlab{a}.
\newblock \showarticletitle{Large-{Scale} {Contrastive} {Language}-{Audio} {Pretraining} with {Feature} {Fusion} and {Keyword}-to-{Caption} {Augmentation}}. In \bibinfo{booktitle}{\emph{{IEEE} {International} {Conference} on {Acoustics}, {Speech} and {Signal} {Processing} ({ICASSP})}}. \bibinfo{pages}{1--5}.
\newblock
\newblock
\shownote{ISSN: 2379-190X}.


\bibitem[Yang et~al\mbox{.}(2023)]%
        {yangAVoiDDF2023}
\bibfield{author}{\bibinfo{person}{Wenyuan Yang}, \bibinfo{person}{Xiaoyu Zhou}, \bibinfo{person}{Zhikai Chen}, \bibinfo{person}{Bofei Guo}, \bibinfo{person}{Zhongjie Ba}, \bibinfo{person}{Zhihua Xia}, \bibinfo{person}{Xiaochun Cao}, {and} \bibinfo{person}{Kui Ren}.} \bibinfo{year}{2023}\natexlab{}.
\newblock \showarticletitle{{AVoiD}-{DF}: {Audio}-{Visual} {Joint} {Learning} for {Detecting} {Deepfake}}.
\newblock \bibinfo{journal}{\emph{IEEE Transactions on Information Forensics and Security}}  \bibinfo{volume}{18} (\bibinfo{year}{2023}), \bibinfo{pages}{2015--2029}.
\newblock
\showISSN{1556-6021}


\bibitem[Yang et~al\mbox{.}(2019)]%
        {yangExposing2019}
\bibfield{author}{\bibinfo{person}{Xin Yang}, \bibinfo{person}{Yuezun Li}, {and} \bibinfo{person}{Siwei Lyu}.} \bibinfo{year}{2019}\natexlab{}.
\newblock \showarticletitle{Exposing {Deep} {Fakes} {Using} {Inconsistent} {Head} {Poses}}. In \bibinfo{booktitle}{\emph{{IEEE} {International} {Conference} on {Acoustics}, {Speech} and {Signal} {Processing} ({ICASSP})}}. \bibinfo{pages}{8261--8265}.
\newblock
\newblock
\shownote{ISSN: 2379-190X}.


\bibitem[Yi et~al\mbox{.}(2022)]%
        {yiADD2022}
\bibfield{author}{\bibinfo{person}{Jiangyan Yi}, \bibinfo{person}{Ruibo Fu}, \bibinfo{person}{Jianhua Tao}, \bibinfo{person}{Shuai Nie}, \bibinfo{person}{Haoxin Ma}, \bibinfo{person}{Chenglong Wang}, \bibinfo{person}{Tao Wang}, \bibinfo{person}{Zhengkun Tian}, \bibinfo{person}{Ye Bai}, \bibinfo{person}{Cunhang Fan}, \bibinfo{person}{Shan Liang}, \bibinfo{person}{Shiming Wang}, \bibinfo{person}{Shuai Zhang}, \bibinfo{person}{Xinrui Yan}, \bibinfo{person}{Le Xu}, \bibinfo{person}{Zhengqi Wen}, \bibinfo{person}{Haizhou Li}, \bibinfo{person}{Zheng Lian}, {and} \bibinfo{person}{Bin Liu}.} \bibinfo{year}{2022}\natexlab{}.
\newblock \bibinfo{title}{{ADD} 2022: the {First} {Audio} {Deep} {Synthesis} {Detection} {Challenge}}.
\newblock
\newblock
\newblock
\shownote{arXiv:2202.08433 [cs, eess]}.


\bibitem[Yu et~al\mbox{.}(2023)]%
        {yuPVASSMDD2023}
\bibfield{author}{\bibinfo{person}{Yang Yu}, \bibinfo{person}{Xiaolong Liu}, \bibinfo{person}{Rongrong Ni}, \bibinfo{person}{Siyuan Yang}, \bibinfo{person}{Yao Zhao}, {and} \bibinfo{person}{Alex~C. Kot}.} \bibinfo{year}{2023}\natexlab{}.
\newblock \showarticletitle{{PVASS}-{MDD}: {Predictive} {Visual}-audio {Alignment} {Self}-supervision for {Multimodal} {Deepfake} {Detection}}.
\newblock \bibinfo{journal}{\emph{IEEE Transactions on Circuits and Systems for Video Technology}} (\bibinfo{year}{2023}), \bibinfo{pages}{1--1}.
\newblock
\showISSN{1558-2205}


\bibitem[Zhang et~al\mbox{.}(2022)]%
        {zhangActionFormer2022}
\bibfield{author}{\bibinfo{person}{Chen-Lin Zhang}, \bibinfo{person}{Jianxin Wu}, {and} \bibinfo{person}{Yin Li}.} \bibinfo{year}{2022}\natexlab{}.
\newblock \showarticletitle{{ActionFormer}: {Localizing} {Moments} of {Actions} with {Transformers}}. In \bibinfo{booktitle}{\emph{Proceedings of the {European} {Conference} on {Computer} {Vision} ({ECCV})}} \emph{(\bibinfo{series}{Lecture {Notes} in {Computer} {Science}})}, \bibfield{editor}{\bibinfo{person}{Shai Avidan}, \bibinfo{person}{Gabriel Brostow}, \bibinfo{person}{Moustapha Cissé}, \bibinfo{person}{Giovanni~Maria Farinella}, {and} \bibinfo{person}{Tal Hassner}} (Eds.). \bibinfo{publisher}{Springer Nature Switzerland}, \bibinfo{address}{Cham}, \bibinfo{pages}{492--510}.
\newblock
\showISBNx{978-3-031-19772-7}


\bibitem[Zhang et~al\mbox{.}(2023a)]%
        {zhangVideoLLaMA2023}
\bibfield{author}{\bibinfo{person}{Hang Zhang}, \bibinfo{person}{Xin Li}, {and} \bibinfo{person}{Lidong Bing}.} \bibinfo{year}{2023}\natexlab{a}.
\newblock \bibinfo{title}{Video-{LLaMA}: {An} {Instruction}-tuned {Audio}-{Visual} {Language} {Model} for {Video} {Understanding}}.
\newblock
\newblock
\newblock
\shownote{arXiv:2306.02858 [cs, eess]}.


\bibitem[Zhang et~al\mbox{.}(2023b)]%
        {zhangUMMAFormer2023}
\bibfield{author}{\bibinfo{person}{Rui Zhang}, \bibinfo{person}{Hongxia Wang}, \bibinfo{person}{Mingshan Du}, \bibinfo{person}{Hanqing Liu}, \bibinfo{person}{Yang Zhou}, {and} \bibinfo{person}{Qiang Zeng}.} \bibinfo{year}{2023}\natexlab{b}.
\newblock \showarticletitle{{UMMAFormer}: {A} {Universal} {Multimodal}-adaptive {Transformer} {Framework} for {Temporal} {Forgery} {Localization}}. In \bibinfo{booktitle}{\emph{Proceedings of the 31st {ACM} {International} {Conference} on {Multimedia}}} \emph{(\bibinfo{series}{{MM} '23})}. \bibinfo{publisher}{Association for Computing Machinery}, \bibinfo{address}{New York, NY, USA}, \bibinfo{pages}{8749--8759}.
\newblock
\showISBNx{9798400701085}


\bibitem[Zhao et~al\mbox{.}(2017)]%
        {zhaoTemporal2017}
\bibfield{author}{\bibinfo{person}{Yue Zhao}, \bibinfo{person}{Yuanjun Xiong}, \bibinfo{person}{Limin Wang}, \bibinfo{person}{Zhirong Wu}, \bibinfo{person}{Xiaoou Tang}, {and} \bibinfo{person}{Dahua Lin}.} \bibinfo{year}{2017}\natexlab{}.
\newblock \showarticletitle{Temporal {Action} {Detection} {With} {Structured} {Segment} {Networks}}. In \bibinfo{booktitle}{\emph{Proceedings of the {IEEE} {International} {Conference} on {Computer} {Vision}}}. \bibinfo{pages}{2914--2923}.
\newblock


\bibitem[Zhou et~al\mbox{.}(2021)]%
        {zhouFace2021}
\bibfield{author}{\bibinfo{person}{Tianfei Zhou}, \bibinfo{person}{Wenguan Wang}, \bibinfo{person}{Zhiyuan Liang}, {and} \bibinfo{person}{Jianbing Shen}.} \bibinfo{year}{2021}\natexlab{}.
\newblock \showarticletitle{Face {Forensics} in the {Wild}}. In \bibinfo{booktitle}{\emph{Proceedings of the {IEEE}/{CVF} {Conference} on {Computer} {Vision} and {Pattern} {Recognition}}}. \bibinfo{pages}{5778--5788}.
\newblock


\bibitem[Zi et~al\mbox{.}(2020)]%
        {ziWildDeepfake2020}
\bibfield{author}{\bibinfo{person}{Bojia Zi}, \bibinfo{person}{Minghao Chang}, \bibinfo{person}{Jingjing Chen}, \bibinfo{person}{Xingjun Ma}, {and} \bibinfo{person}{Yu-Gang Jiang}.} \bibinfo{year}{2020}\natexlab{}.
\newblock \showarticletitle{{WildDeepfake}: {A} {Challenging} {Real}-{World} {Dataset} for {Deepfake} {Detection}}. In \bibinfo{booktitle}{\emph{Proceedings of the 28th {ACM} {International} {Conference} on {Multimedia}}} \emph{(\bibinfo{series}{{MM} '20})}. \bibinfo{publisher}{Association for Computing Machinery}, \bibinfo{address}{New York, NY, USA}, \bibinfo{pages}{2382--2390}.
\newblock
\showISBNx{978-1-4503-7988-5}


\end{thebibliography}

\appendix
\clearpage

{
\newpage
   \twocolumn[
    \centering
    \Large
    \textbf{AV-Deepfake1M: A Large-Scale LLM-Driven Audio-Visual Deepfake Dataset}\\
    \vspace{0.5em}Supplementary Material \\
    \vspace{1.0em}
   ] 
}
   
\section{Transcript Manipulation}
\begin{figure}[t]
\centering
\includegraphics[width=\linewidth]{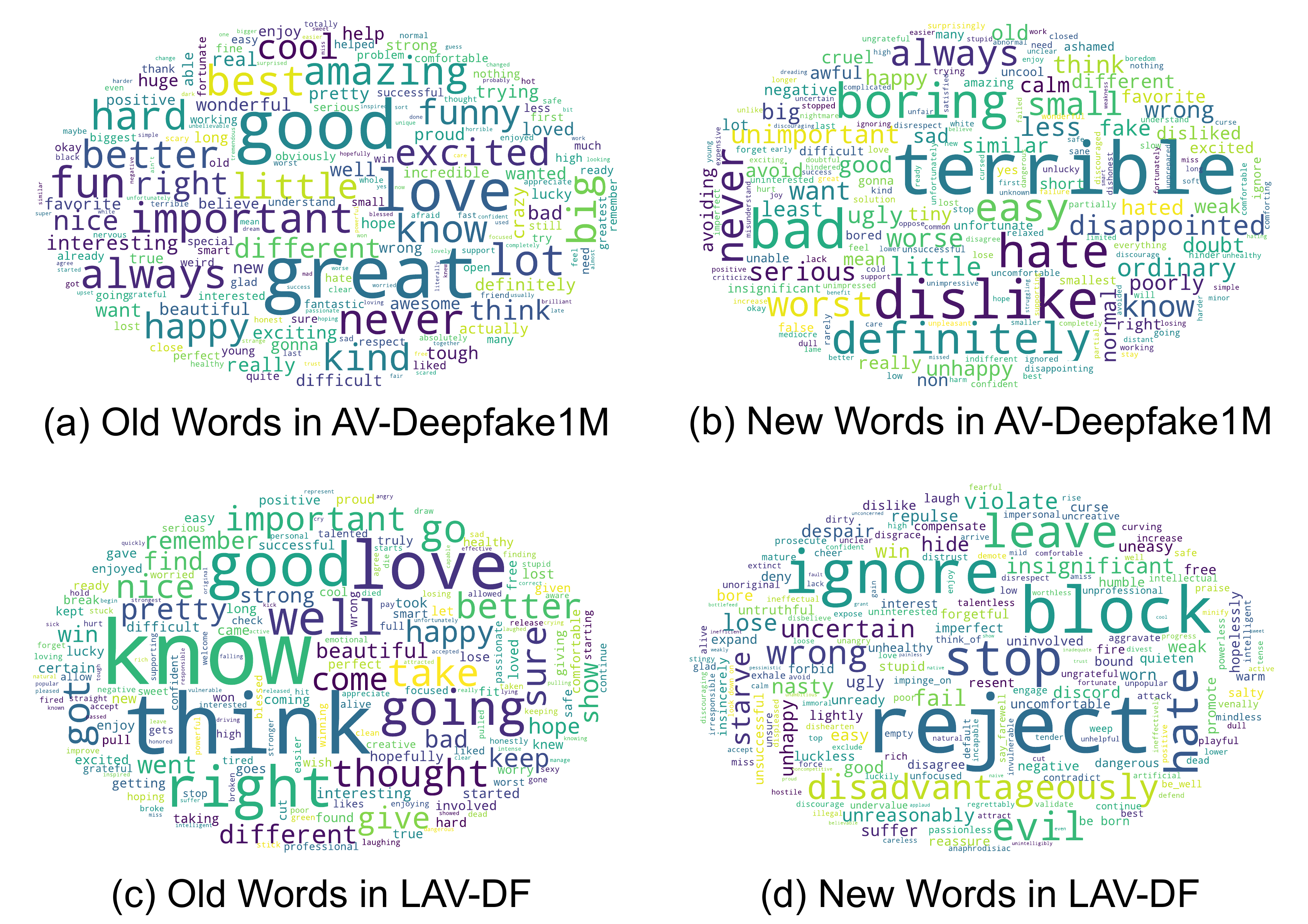}
\caption{\textbf{Qualitative comparison of transcript modifications in \datasetabbr{} and LAV-DF.} \textmd{(a) The old words before the manipulations in \datasetabbr{}. (b) The new words after the LLM-driven manipulations in \datasetabbr{}. (c) The old words before manipulations in LAV-DF. (d) The new words after the rule-based manipulations in LAV-DF.}}
\label{fig:wordcloud}
\end{figure}

In addition to the quantitative comparison of transcript modifications in AV-Deepfake1M and LAV-DF~\cite{caiYou2022} (see Section~\ref{sec:transcript_manipulation}), here we also present a qualitative one.
Figure~\ref{fig:wordcloud} illustrates word clouds for \texttt{old\_word}(s) and \texttt{new\_word}(s) for both datasets.
A comparison between the new words generated by the rule-based strategy utilized in LAV-DF and our LLM-driven generation further demonstrates that the latter results in more natural and diverse transcript manipulations.

\section{Human Quality Assessment}
\begin{figure}[t]
\centering
\includegraphics[width=\linewidth]{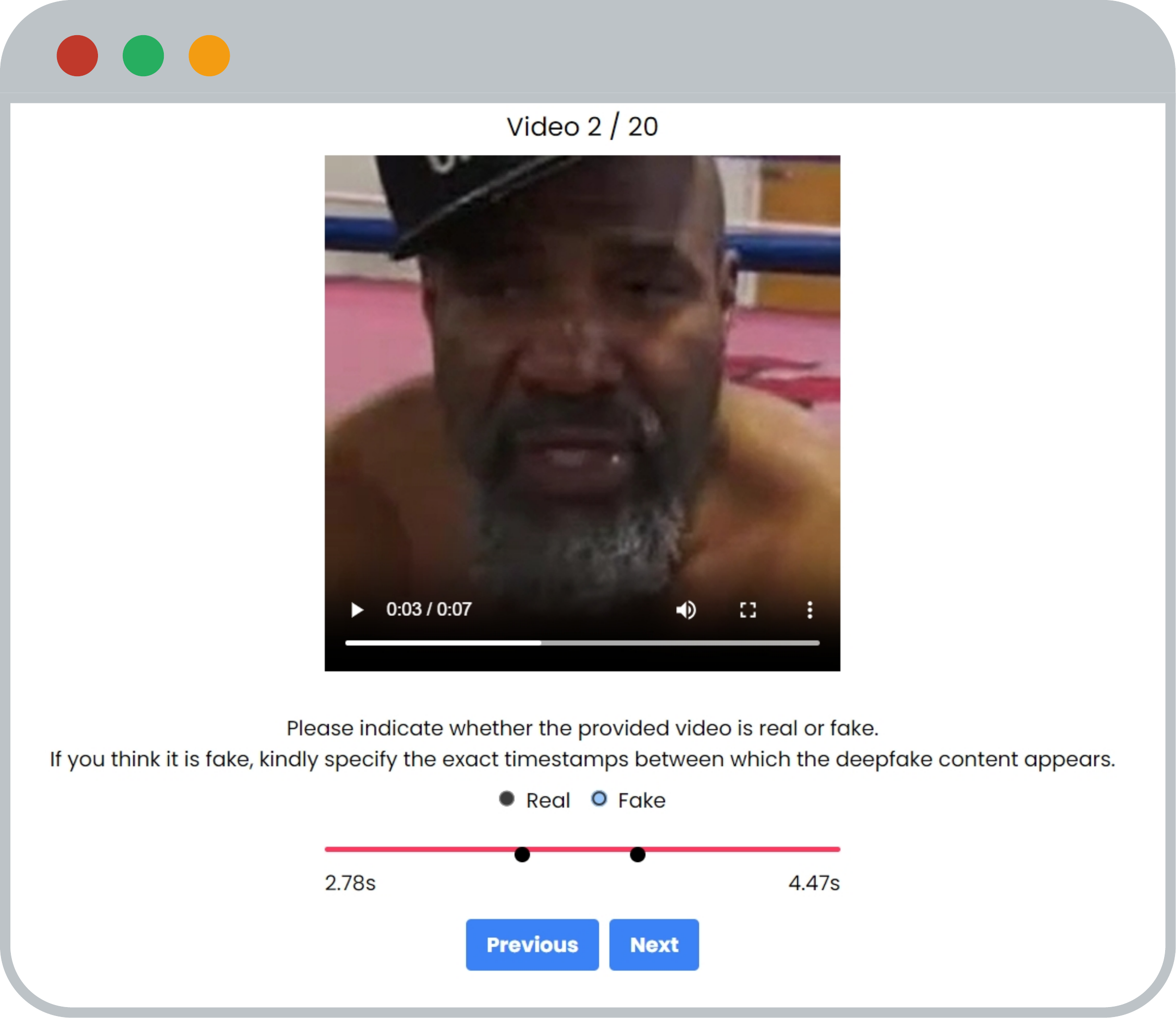}
\caption{\textbf{Screenshot of the user study interface.} \textmd{On the top is the video with audio, the middle is the textual description of the task, and the bottom is the participant's controls to 1) Select whether the video is \emph{real} or \emph{fake} and 2) If the participant selects \emph{fake}, use a slider to specify the begin and end of the fake segment.}}
\label{fig:user_study_screenshot}
\end{figure}

\begin{figure*}[t]
\centering
\includegraphics[width=0.49\linewidth]{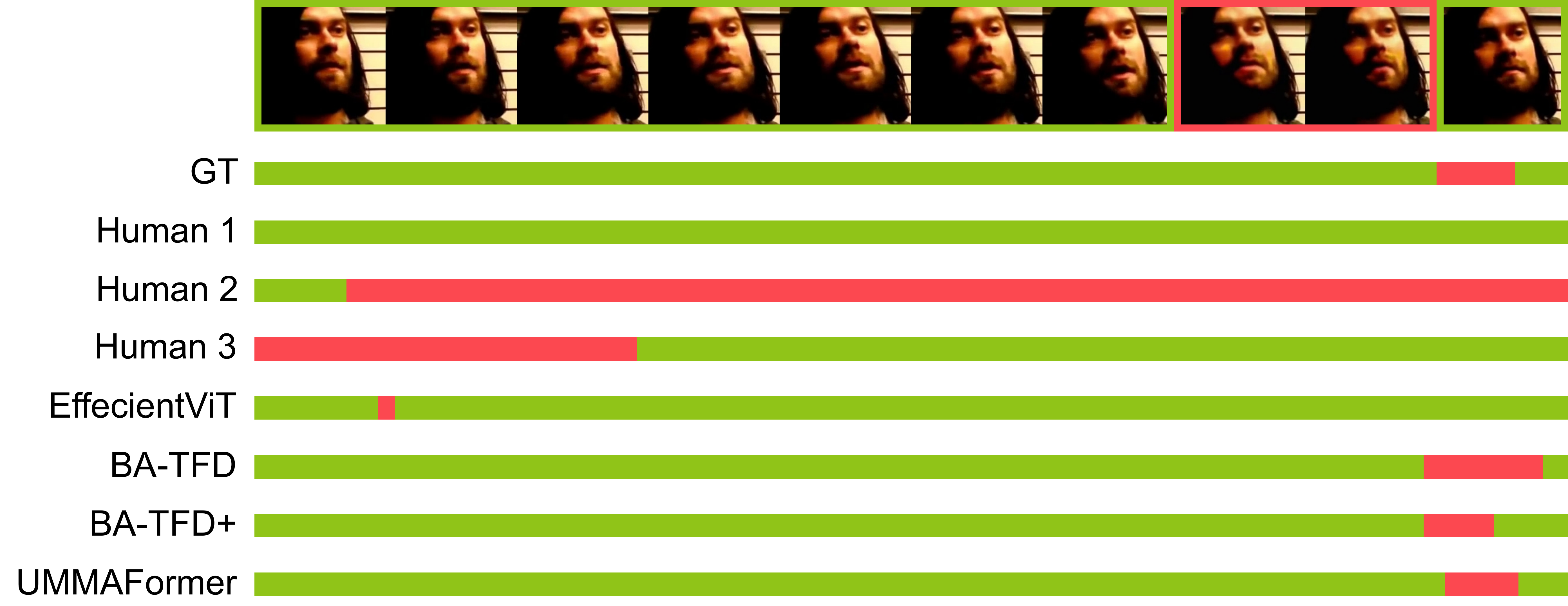}
\includegraphics[width=0.49\linewidth]{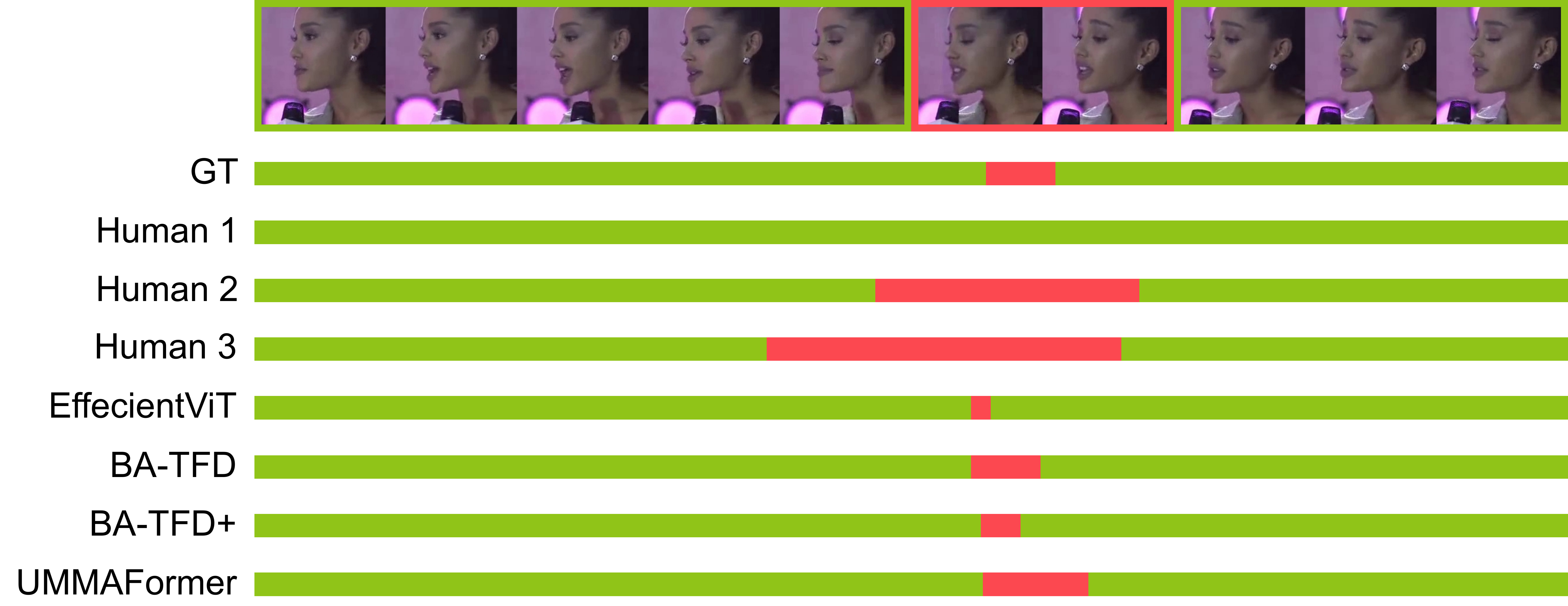}
\includegraphics[width=0.49\linewidth]{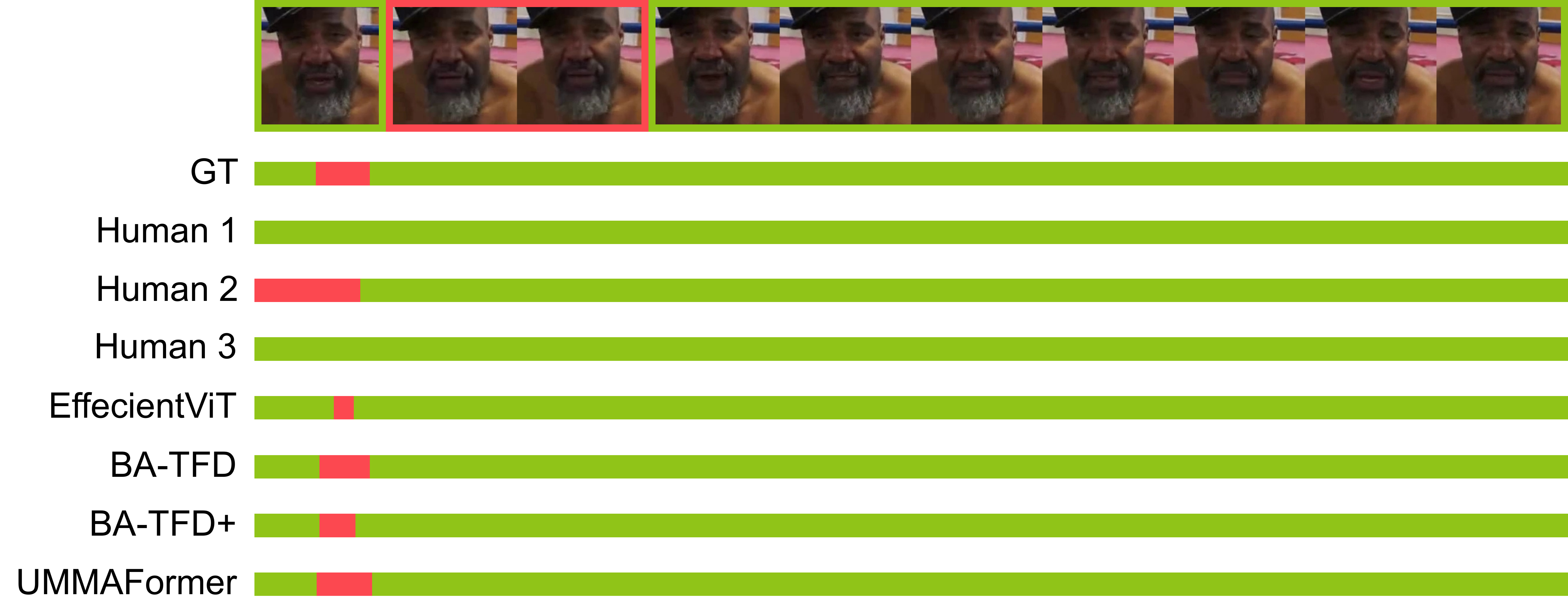}
\includegraphics[width=0.49\linewidth]{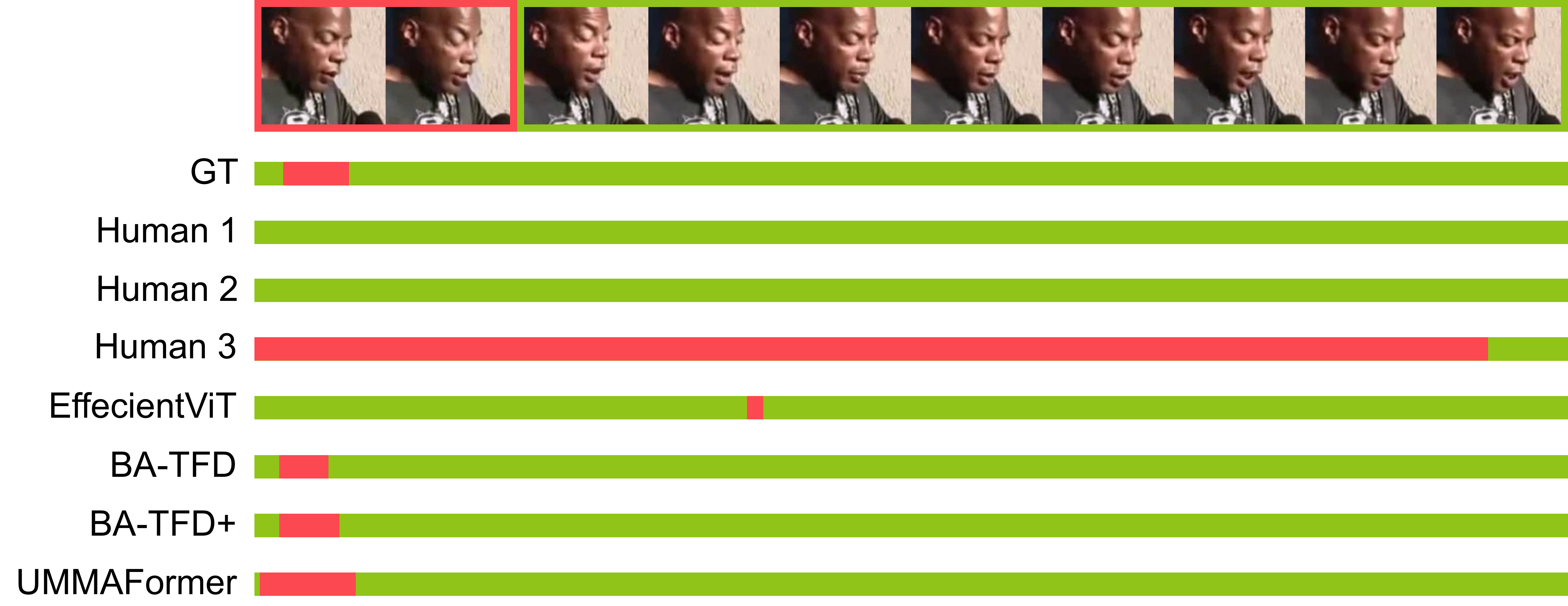}
\includegraphics[width=0.49\linewidth]{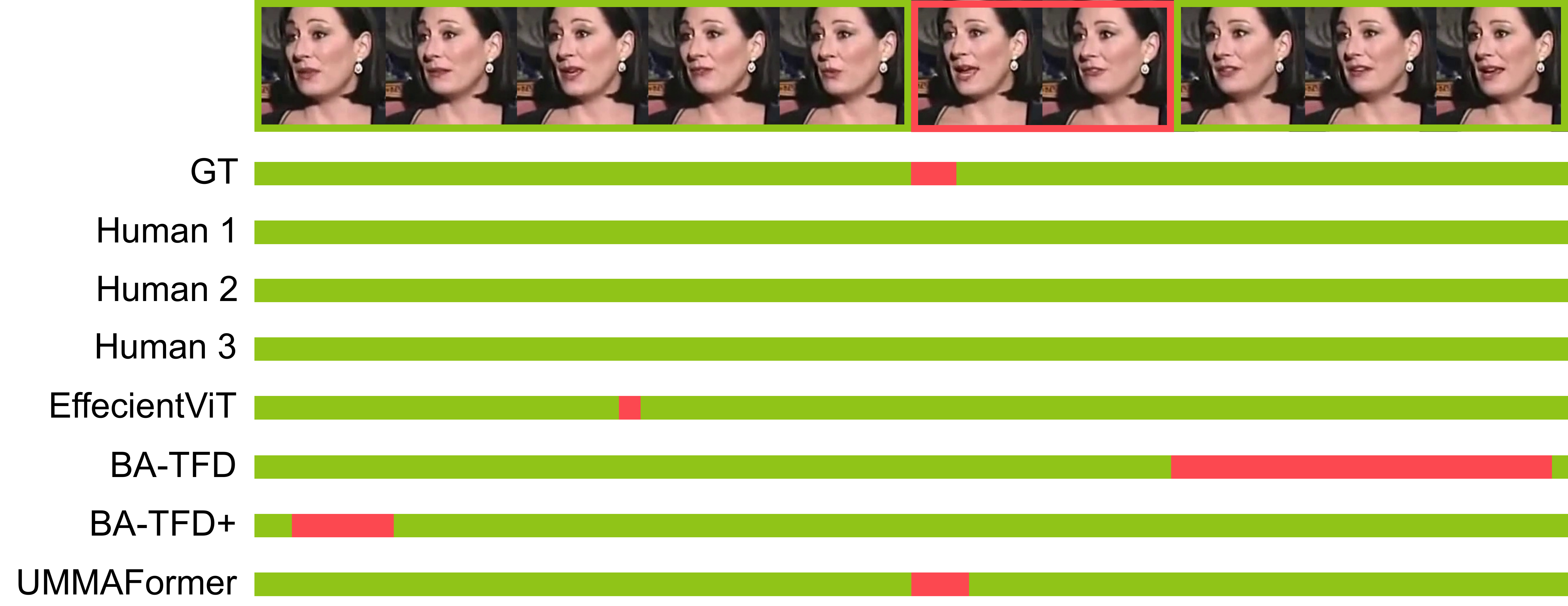}
\includegraphics[width=0.49\linewidth]{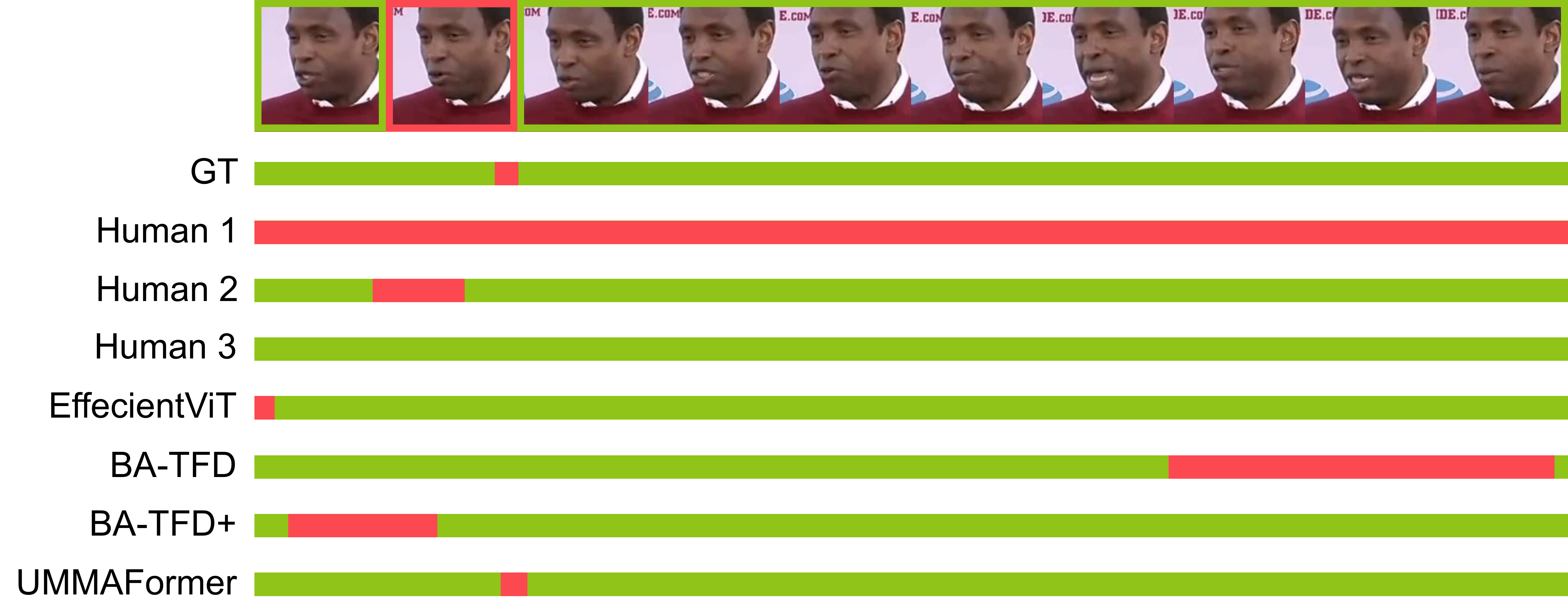}
\includegraphics[width=0.49\linewidth]{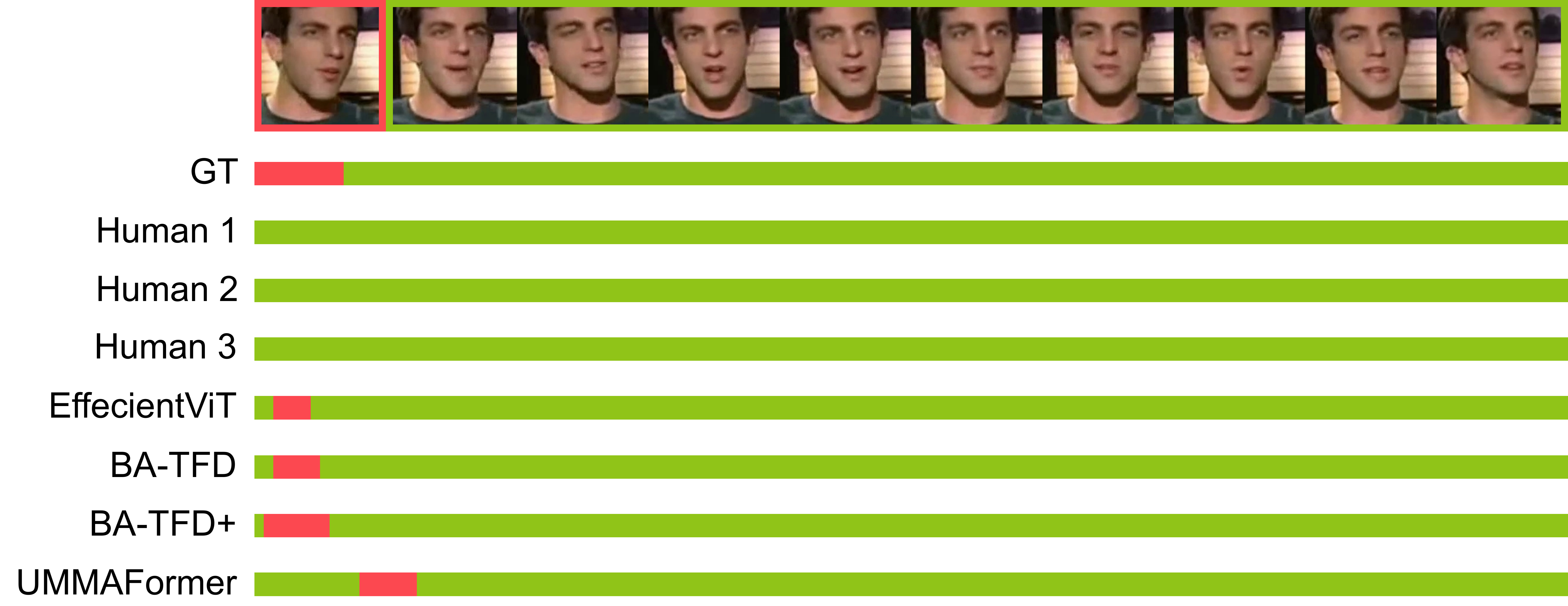}
\includegraphics[width=0.49\linewidth]{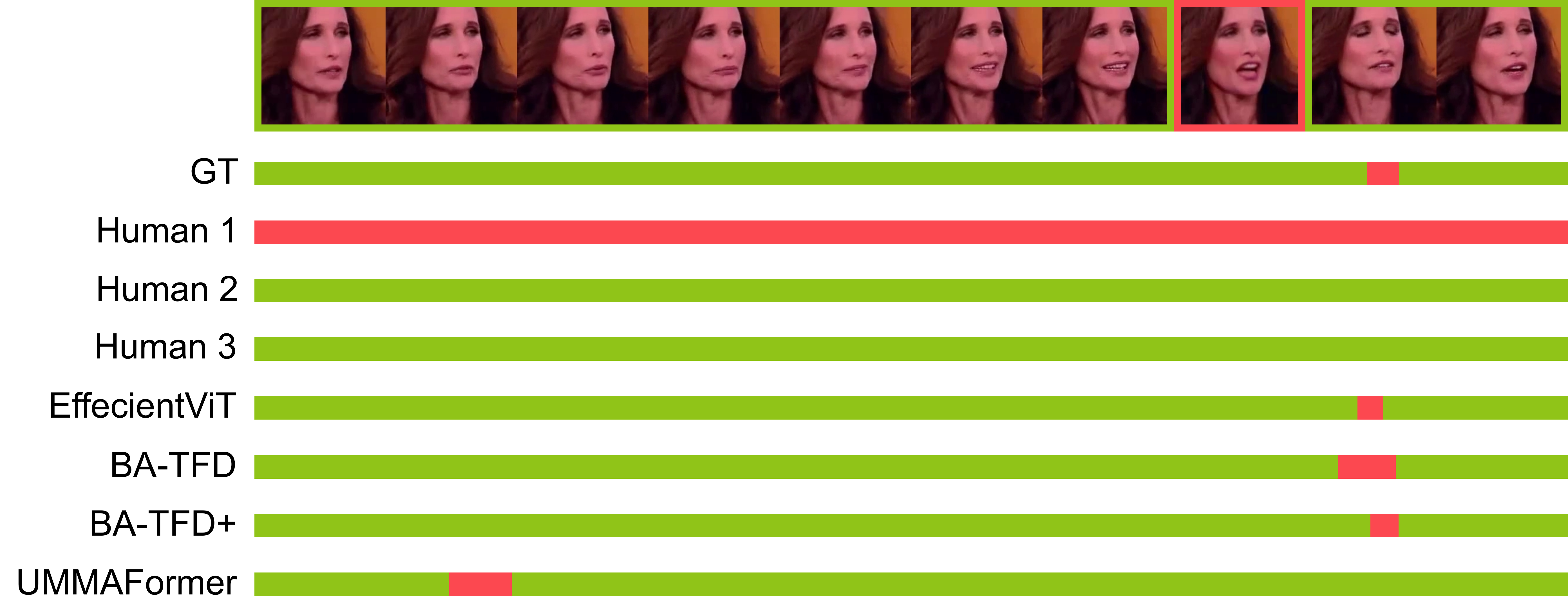}
\caption{\textbf{Examples of user study results and comparison with the state-in-the-art in temporal deepfake localization.} \textmd{\textcolor{ForestGreen}{Green} color represents \emph{real} segments and \textcolor{red}{red} color represents \emph{fake} segments. GT: Ground truth.}}
\label{fig:user_study_example}
\end{figure*}

Here we provide further details on the user study (see Section~\ref{sec:human_quality_assessment}) that aims to evaluate humans' performance in detecting the highly realistic deepfake content in \datasetabbr{}.

\subsection{Data}
The data used in the user study are 200 videos randomly sampled from the \emph{test} set of \datasetabbr{} and LAV-DF~\cite{caiYou2022},
with the aim to maximize the number of unique identities. Please note that the user study setup ensures each participant cannot see a duplicated identity.
The videos include 50 real videos from \datasetabbr{}, 50 fake videos from \datasetabbr{}, 50 real videos from LAV-DF, and 50 fake videos from LAV-DF. For fair comparison with LAV-DF, the fake videos contain only one audio-visual \emph{replacement} (see Section~\ref{sec:Dataset}).

\begin{table*}[t]
\centering
\caption{\textbf{User study results compared with the state-in-the-art in temporal deepfake localization.}}
\scalebox{0.88}{
\begin{tabular}{l|cccc|cccc}
\toprule[0.4mm]
\rowcolor{mygray} \textbf{Dataset} & \multicolumn{4} {c}{\textbf{LAV-DF}~\cite{caiYou2022}} & \multicolumn{4} {c}{\textbf{AV-Deepfake1M}} \\
\rowcolor{mygray} \textbf{Method} & \textbf{Acc.} & \textbf{AP@0.1} & \textbf{AP@0.5} & \textbf{AR@1} & \textbf{Acc.} & \textbf{AP@0.1} & \textbf{AP@0.5} & \textbf{AR@1} \\ \hline \hline
Xception~\cite{cholletXception2017} & 96.00 & 69.33 & 41.75 & 30.40 & 77.00 & 58.78 & 24.26 & 12.20 \\
BA-TFD~\cite{caiYou2022} & - & 95.37 & 80.33 & 66.44 & - & 59.69 & 44.87 & 21.27 \\
BA-TFD+~\cite{caiGlitch2023} & - & 98.00 & 98.00 & 87.60 & - & 65.44 & 51.41 & 23.26 \\
UMMAFormer~\cite{zhangUMMAFormer2023} & - & 98.00 & 98.00 & 97.80 & - & 69.77 & 53.72 & 38.39 \\ \hline
Human & 84.03 & 36.80 & 14.17 & 10.04 & 68.64 & 15.32 & 01.92 & 02.54 \\
\bottomrule[0.4mm]
\end{tabular}}
\label{tab:user_study_comparison}
\end{table*}

\begin{figure*}[t]
\centering
\includegraphics[width=\linewidth]{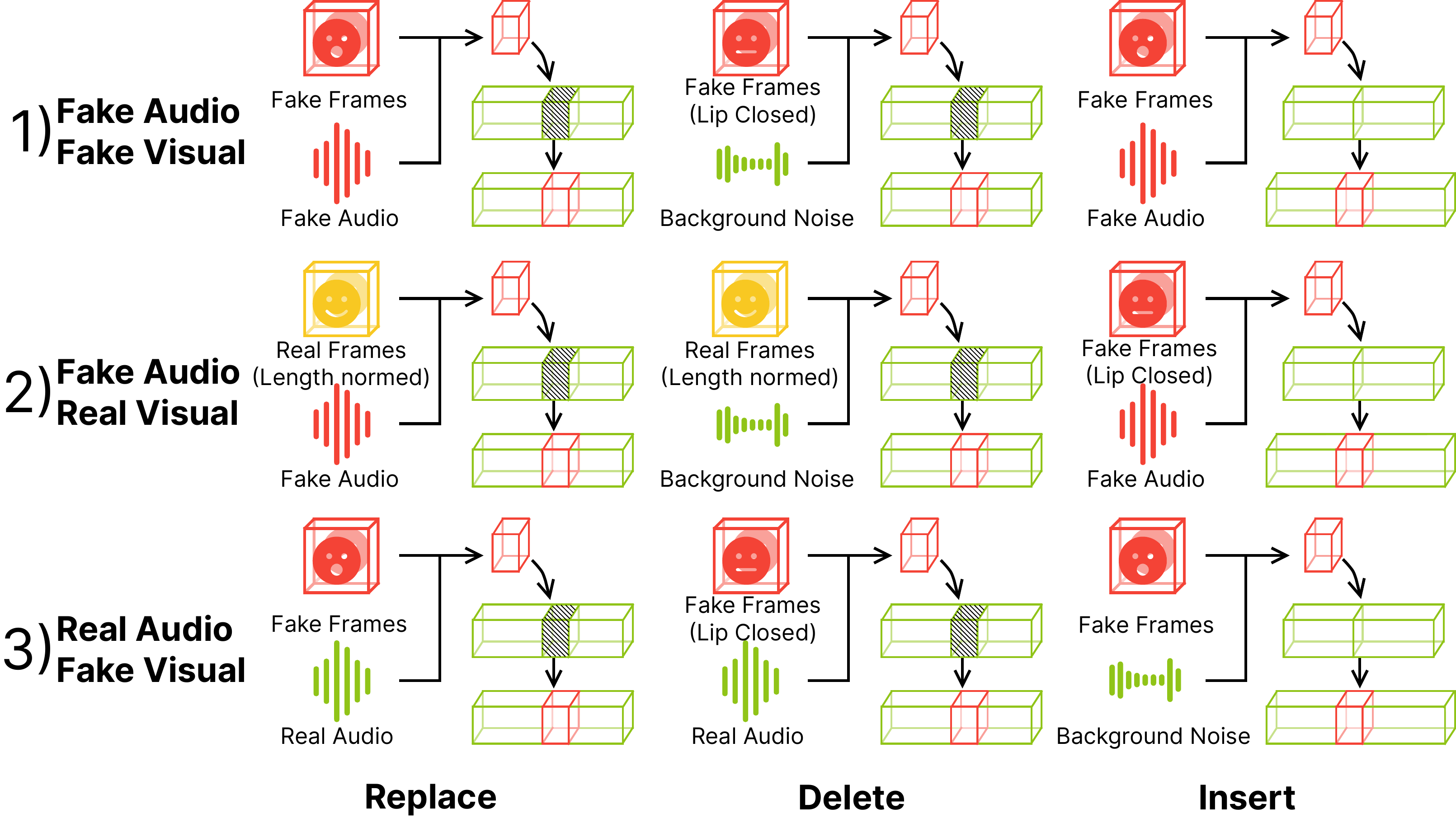}
\caption{\textbf{Details of the audio-visual content generation.} \textmd{Here, we show the audio-visual content manipulation strategy in three setups i.e. fake audio fake video, fake audio real video and real audio fake video. We believe that these three variations of fake content generation add more challenge in the temporal localization task.}}
\label{fig:modalities}
\end{figure*}

\begin{figure*}[t]
\centering
\includegraphics[width=\linewidth]{image/label_access_level2.pdf}
\caption{\textbf{Complete details on the label access for training.} \textmd{\textcolor{ForestGreen}{Green} color represents the \emph{real} and \textcolor{red}{red} color represents \emph{fake} content. The top row represents the original \emph{frame-level} labels in a video. The middle row represents the segment- and video-level labels based on whether the segment/video contains any \emph{fake} frames. For fair comparison across different methods, the bottom row represents the mapped segment- and video-level labels to frame-level labels.}}
\label{fig:label_access_level}
\end{figure*}

\subsection{Participants}
We randomly group the participants into 10 groups where each group evaluates 10\% of the videos (\ie, 20 videos including 5 real videos from \datasetabbr{}, 5 fake videos from \datasetabbr{}, 5 real videos from LAV-DF, and 5 fake videos from LAV-DF).
We utilize a random non-overlapping selection of videos for each participant, meaning that each participant evaluates videos for 20 out of the 200 videos.
After watching each video, the participants first answer whether the video is \emph{real} or \emph{fake}, and if they think the video is \emph{fake}, the participants can choose the start and end timestamps for the fake segment.
A screenshot of the developed user study interface based on the React\footnote{\url{https://react.dev/}} framework is shown in Figure~\ref{fig:user_study_screenshot}.

\subsection{Evaluation and Analysis}
Among the 25 participants that took part in the user study, the binary deepfake detection/classification accuracy is 64.84\% for \datasetabbr{}.
This low performance indicates that the deepfake content in \datasetabbr{} is very challenging for humans to detect.
A similar pattern is observed for the temporal localization of fake segments.
Similarly to Table~\ref{tab:user_study}, here we report and compare average precision (AP) and average recall (AR) scores in Table~\ref{tab:user_study_comparison} and extend that comparison with the state-of-the-art methods using the same subset of videos.
The AP score for 0.5 IoU is 01.92.
Thus, we reduced the AP threshold to 0.1 IoU, improving the AP score to 15.32.
Figure~\ref{fig:user_study_example} illustrates a similar qualitative comparison.
The low human performance in each aspect indicates that to detect highly realistic deepfake content, we need more sophisticated detection and localization methods.

Considering LAV-DF~\cite{caiYou2022}, we observed similar patterns - human performance is lower than the state-of-the-art detection and localization methods. Comparing the human performance between \datasetabbr{} (Acc. 68.64, AP@0.1 15.32) and LAV-DF (Acc. 84.03, AP@0.1 36.80), we find that \datasetabbr{} is more challenging than LAV-DF for humans.

\section{Audio and Video Generation}
Here we provide complete details on the manipulations in \datasetabbr{} (see Section~\ref{sec:Dataset}).
Figure~\ref{fig:modalities} provides visualizations corresponding to each of the three modifications and the resulting deepfake content. Please note that for example for \textbf{Fake Audio} and \textbf{Real Visual} in the cases of \emph{deletion} and \emph{insertion}, there are slight modifications in the visual signal as well. The reason we regard the visual signal as \emph{real} is the fact that words were not \emph{inserted} or \emph{deleted} in that modality. Similarly for \textbf{Real Audio} and \textbf{Fake Visual}.

\section{Label Access For Training}
Figure~\ref{fig:label_access_level} provides complete details on the label access during training (see Section~\ref{sec:audio-visual_deepfake_detection}).

\begin{itemize}
\item{In the \emph{frame-level} configuration, the models are trained using the ground truth labels for each frame in the video.}
\item{In the \emph{segment-level} configuration, if the segment contains any \emph{fake} frames, it is labelled as \emph{fake} otherwise it is labelled as \emph{real}.
For the segment-based methods MARLIN~\cite{caiMARLIN2023} and MDS~\cite{chughNot2020}, we used the \emph{segment-level} labels during training.
For a fair comparison when training the frame-based methods Meso4~\cite{afcharMesoNet2018} and MesoInception4~\cite{afcharMesoNet2018} we mapped the \emph{segment-level} labels to \emph{frame-level}.}
\item{In the \emph{video-level} configuration, if the video contains any \emph{fake} frames, it is labelled as \emph{fake} otherwise it is labelled as \emph{real}.
Similarly to the \emph{segment-level} configuration, for a fair comparison when training the frame-based methods Meso4~\cite{afcharMesoNet2018} and MesoInception4~\cite{afcharMesoNet2018} we mapped the \emph{video-level} labels to \emph{frame-level}.}
\end{itemize}

\end{document}